\documentclass[acmsmall]{acmart}
\usepackage[utf8]{inputenc}
\usepackage[T1]{fontenc}

\AtBeginDocument{%
  }

\setcopyright{acmlicensed}
\copyrightyear{2018}
\acmYear{2018}
\acmDOI{XXXXXXX.XXXXXXX}

\acmJournal{JACM}
\acmVolume{37}
\acmNumber{4}
\acmArticle{111}
\acmMonth{8}

\usepackage[most]{tcolorbox}

\usepackage{multirow}
\usepackage{mdframed}
\usepackage{xcolor}
\usepackage{soul}  
\usepackage{tikz}
\usepackage{subfigure}
\usetikzlibrary{arrows.meta, positioning}
\usepackage[utf8]{inputenc}
\usepackage[ruled,vlined,linesnumbered]{algorithm2e}

\usepackage{amsmath,amssymb}
\usepackage{tabularx}
\usepackage{array}
\usepackage{hyperref}

\usepackage{todonotes}
\newcounter{todocounter}
\usepackage{soul,xcolor}
\sethlcolor{yellow!20} 

\newcommand{\added}[1]{\textcolor{blue}{#1}}

\newmdenv[
  linecolor=gray!70,
  backgroundcolor=gray!10,
  roundcorner=5pt,
  innertopmargin=1em,
  innerbottommargin=1em,
  innerleftmargin=1em,
  innerrightmargin=1em
]{MyFrame}

\sethlcolor{yellow!20}               

\tcbset{
  algobox/.style={
    colback=blue!5!white,
    colframe=blue!75!black,
    sharp corners,
    boxrule=1pt,
    left=1em, right=1em, top=1em, bottom=1em,
    title=Autophagy Pipeline
  }
}

\begin{document}

\title{Learning by Surprise: Adaptive Mitigation of Model Collapse in Large Language Models}

\author[Daniele Gambetta]{Daniele Gambetta}
\affiliation{%
  \institution{University of Pisa}
  \city{Pisa}
  \country{Italy}
}
\affiliation{%
  \institution{ISTI-CNR}
  \city{Pisa}
  \country{Italy}
}
\email{daniele.gambetta@phd.unipi.it}

\author[Gizem Gezici]{Gizem Gezici}
\affiliation{%
  \institution{Scuola Normale Superiore}
  \city{Pisa}
  \country{Italy}
}
\email{gizem.gezici@sns.it}

\author[Fosca Giannotti]{Fosca Giannotti}
\affiliation{%
  \institution{Scuola Normale Superiore}
  \city{Pisa}
  \country{Italy}
}
\email{fosca.giannotti@sns.it}

\author[Dino Pedreschi]{Dino Pedreschi}
\affiliation{%
  \institution{University of Pisa}
  \city{Pisa}
  \country{Italy}
}
\email{dino.pedreschi@unipi.it}

\author[Alistair Knott]{Alistair Knott}
\affiliation{%
  \institution{Victoria University of Wellington}
  \city{Wellington}
  \country{New Zealand}
}
\email{ali.knott@vuw.ac.nz}

\author[Luca Pappalardo]{Luca Pappalardo}
\affiliation{%
  \institution{ISTI-CNR}
  \city{Pisa}
  \country{Italy}
}
\affiliation{%
  \institution{Scuola Normale Superiore}
  \city{Pisa}
  \country{Italy}
}
\email{luca.pappalardo@isti.cnr.it}

\renewcommand{\shortauthors}{Gambetta et al.}



\begin{abstract}
As AI-generated content increasingly populates the web, generative AI models are at growing risk of being trained on their own outputs, a process known as AI autophagy. 
This feedback loop has been shown to induce model collapse, typically characterized by a loss of diversity in generated content. However, existing work offers a limited understanding of this phenomenon and relies on mitigation strategies that assume access to human-authored data.
In this paper, we conduct extensive simulations across multiple datasets and LLMs to address key gaps in the study of model collapse. First, we introduce model-intrinsic measures based on next-token probability distributions, showing that model collapse corresponds to an increasing concentration of probability mass on a small set of tokens. 
Second, we demonstrate that model collapse is also associated with a loss of common sense, as measured by a decline in commonsense inference accuracy. 
Third, we identify perplexity (a measure of model ``surprise'') as a key driver of collapse: fine-tuning on the least "surprising" documents leads to more severe degeneration.
Building on this insight, we propose a perplexity-based filtering strategy that prioritizes high-surprise documents during fine-tuning. Unlike existing approaches, our method does not require distinguishing between human-authored and AI-generated content. Across datasets and LLM families, this strategy consistently mitigates model collapse, achieving performance comparable to, and in some cases better than, human-data baselines, while substantially reducing the concentration of next-token probabilities.
Overall, our results provide a unified, model-centric understanding of model collapse and suggest practical, scalable strategies for training generative AI systems in increasingly synthetic environments.
\end{abstract}



\received{20 February 2007}
\received[revised]{12 March 2009}
\received[accepted]{5 June 2009}

\maketitle

\section{Introduction}
\label{sec:intro}

Generative AI has demonstrated remarkable advancements in recent years, particularly in conversational systems such as ChatGPT, Claude, Google Gemini, and Meta AI. 
These applications have been widely adopted and play a central role in content generation across several online platforms ~\cite{ghassemi2023chatgpt}.

Large Generative Models (LGMs) powering these applications are trained on vast amounts of human-authored content. 
For example, Large Language Models (LLMs) are typically trained on text corpora containing up to two trillion tokens~\cite{touvron2023llama}. 
In comparison, the global pool of high-quality human-authored text is estimated at just 17 trillion tokens, with a modest annual growth rate of 4–5\%~\cite{villalobos2022will}. 
A 2021 report predicts that already by 2025, up to 90\% of internet content could be AI-generated~\cite{europol}, heralding what has been described as the Age of Synthetic Realities~\cite{cardenuto2023age}.
As LGMs' training relies on web-sourced content, there is a rising risk that their own outputs -- the AI-generated content -- will be used to train and fine-tune future models, creating a self-consuming feedback loop~\cite{pedreschi2024human}.

This feedback loop is often referred to as \emph{AI autophagy}, i.e., a process where AI models are recursively fine-tuned on their own outputs~\cite{pappalardo2024survey,pedreschi2024human, xing2025caveats}. 
Recent research shows that AI autophagy leads to a phenomenon known as \emph{model collapse}, characterized by a significant loss of diversity in AI-generated content \cite{shumailov2024ai,alemohammad2023self,guo2023curious, pappalardo2024survey}. 
Our understanding of model collapse is still evolving, with many questions remaining open~\cite{schaeffer2025positionmodelcollapsedoes}. 

In this paper, through extensive simulations across three text datasets and four LLMs, we investigate and address four key gaps in the current understanding of model collapse:

\begin{enumerate}
\item \textbf{Model-intrinsic characterization}: model collapse is typically diagnosed by monitoring metrics computed on the AI-generated content. 
We lack model-intrinsic measures that characterize collapse from the LLM's next-token probability distributions;
\item \textbf{Model collapse as a loss of common sense}: it remains unclear whether model collapse also impairs the LLM’s ability to generate meaningful and commonsensical text.
\item \textbf{Drivers of model collapse:} it remains unclear how specific properties of AI-generated content contribute, drive, and accelerate model collapse;
\item \textbf{Mitigation of model collapse:} the majority of mitigation techniques for model collapse rely on curated, high-quality data subsets and on the assumption that it is possible to distinguish human-authored from AI-generated content. However, as AI-generated content increasingly saturates the web, such reliable source signals become progressively unavailable.
\end{enumerate}

To address (1), we introduce model-intrinsic measures for characterizing model collapse. 
We define an LLM as \emph{collapsed} when its next-token probability distributions become highly concentrated, i.e., they assign a disproportionate share of probability mass to a small subset of tokens. 
We quantify this concentration using the Gini coefficient of the next-token distribution and an indicator capturing whether the maximum token probability is near the value of one.

We address (2) by introducing a measure that captures the probability that an LLM generates a commonsensical continuation of a given text. We show that model collapse is also associated with a loss of common sense, as this measure consistently declines as AI autophagy unfolds.

We address (3) by showing that, given a document, the model’s \emph{surprise} with respect to that document -- measured via \emph{perplexity} -- provides a principled bridge between model collapse and extrinsic properties of data used for model fine-tuning.
The perplexity measure quantifies how predictable a document is under an LLM, allowing us to operationalize dataset characteristics such as diversity, novelty, and redundancy \cite{kovavc2025recursive}. Through this lens, we demonstrate how shifts in these extrinsic properties systematically translate into changes in the LLM’s internal next-token probability distribution, thereby driving and revealing the onset of model collapse.

We then address (4) by introducing a \emph{perplexity-based filtering strategy} that selects high-surprise documents for fine-tuning.
We show that, across datasets and LLM families, this strategy consistently mitigates model collapse, achieving performance comparable to -- and in some metrics better than -- fine-tuning on human-authored data alone, while requiring no knowledge of whether the training documents are human-authored or AI-generated.
In this way, our approach extends prior detector-based and accumulation methods~\cite{drayson2025machine,gerstgrasser2024model} to unlabeled web corpora.

The paper is organized as follows. Section~\ref{sec:related} surveys prior research and positions our work within the current literature.
Section~\ref{sec:definitions} formalizes the concepts of AI autophagy and model collapse.
Section~\ref{sec:framework} details our AI autophagy simulation framework and Section~\ref{sec:measures} formalizes the measures used to characterize model collapse. 
In Section~\ref{sec:settings}, we describe our experimental settings and in Section~\ref{sec:initial_experiments} we characterize model collapse using our model-intrinsic measures. 
Section~\ref{sec:mitigation} evaluates our perplexity-based data filtering strategy and shows it matches human-data performance without requiring labels to distinguish between human-authored and AI-generated documents.
Finally, Section~\ref{sec:conclusions} concludes the paper, discussing limitations and outlining directions for future research.


\section{Related Work}
\label{sec:related}

Research on AI autophagy and model collapse relies on simulations because large-scale, longitudinal empirical studies are hindered by the difficulty of tracking the many evolving versions of generative models deployed on online platforms, especially as they are frequently fine-tuned and updated over time~\cite{pappalardo2024survey,pedreschi2024human}.
These simulations show that model collapse occurs in various AI models including LGMs, Variational Autoencoders (VAEs) and Gaussian Mixture Models (GMMs) ~\cite{shumailov2024ai,guo2023curious,briesch2023large,dohmatob2024tail,alemohammad2023self,martinez2023combining,martinez2023towards,dohmatob2024model,hataya2023will,bohacek2023nepotistically}.

Shumailov et al.~\cite{shumailov2024ai} introduce the concepts of AI autophagy and model collapse. 
Several studies examine how different mixtures of human-authored and AI-generated data affect model collapse. 
Alemohammad et al.~\cite{alemohammad2023self} show that human-authored data can delay model collapse but not prevent it.
Martínez et al.~\cite{martinez2023towards} train a generative model on a 50/50 mix of human-authored and AI-generated images, observing that, over time, the model's outputs become more similar to human-authored images but lose diversity.
Briesch et al.~\cite{briesch2023large} compare a fully synthetic cycle where training data is entirely replaced each generation; incremental and balanced cycles; and an expanding cycle that continuously adds new data. 
They find that 
only the expanding cycle can preserve diversity over longer horizons, maintaining it for up to 50 generations. 
Other studies examine the inevitability of model collapse when models are trained on AI-generated data and estimate the minimum proportion of human-authored data required to address it~\cite{bertrand2024stability, seddik2024bad}.

Guo et al.~\cite{guo2023curious} introduce metrics to measure lexical, syntactic, and semantic diversity as AI autophagy progresses.
Studying three use cases -- news summarization, scientific abstract generation, and story generation -- they find that the decline in linguistic diversity is more pronounced in tasks that demand greater creativity.
The effect of simulation parameters on model collapse has also been investigated: Herel and Mikolov~\cite{herel2024collapse} show that increasing the learning rate accelerates the onset of model collapse, while Suresh et al.~\cite{suresh2024ratemodelcollapserecursive} find that the time it takes for a model to forget a word is linearly related to that word’s frequency in the original training corpus.

Several studies investigate model collapse in the text-to-image domain. 
Martínez et al.~\cite{martinez2023combining} show that augmenting training data with AI-generated images leads to a progressive decline in image quality across generations. 
Hataya et al.~\cite{hataya2023will} study synthetic data contamination and find that performance deteriorates as the proportion of AI-generated images increases. They also propose a method that uses a masked autoencoder to detect AI-generated images.
Bohacek and Farid~\cite{bohacek2023nepotistically} show that model collapse emerges regardless of the proportion of human-authored versus AI-generated images used for retraining, but can be reversed by fine-tuning exclusively on human-authored images. 
\citet{dohmatob2024model, dohmatob2024tail} extend the analysis to text generation, showing that while mixed training initially improves performance, it eventually leads to model collapse.

Another strand of research focuses on strategies to mitigate model collapse. 
Some works propose specific accumulation-based AI autophagy to slow model collapse~\cite{gerstgrasser2024model, briesch2023large}. Others focus on verifying AI-generated data before reusing it. 
\citet{feng2024beyond} show that synthetic data can aid fine-tuning when paired with simple verification against ground truth, helping prevent model collapse. 
Fu et al.~\cite{fu2025theoreticalperspectivepreventmodel} study how the proportion of human-authored versus AI-generated data affects the stability of model collapse, while Zhu et al.~\cite{zhu2024synthesizetextdatamodel} explore how to synthesize data in ways that avoid model collapse, showing that simple token-level edits to human-authored text (creating semi-synthetic data) can be effective.
Drayson et al.~\cite{drayson2025machine} propose incorporating AI-generated text detection mechanisms as a preventive strategy, while \citet{kovavc2025recursive} identify document properties modulating model collapse severity through regression analysis on data clusters.



\paragraph{Position of our work.} 
Existing work mitigates model collapse by preserving or incorporating human-authored data \cite{alemohammad2023self, briesch2023large, martinez2023towards}, leveraging AI-detection mechanisms \cite{drayson2025machine}, and controlling the accumulation of synthetic content across iterations \cite{gerstgrasser2024model}. However, these approaches assume that human- and AI-generated text can be reliably distinguished, an assumption that is becoming increasingly untenable as AI-generated and hybrid content proliferate. We therefore ask: given a set of text documents, what properties make them more likely to trigger model collapse during fine-tuning? This question is motivated by the observation that repeated fine-tuning on the same human-authored data can also induce collapse-like effects.

\section{The Model Collapse problem}
\label{sec:definitions}

In this section, we formalize two core concepts: AI autophagy and model collapse. We assume the generative model is a Large Language Model (LLM) -- foundation models focused on understanding and generating human language \cite{paass2023foundation} -- but the definitions naturally extend to other types of models.

Table \ref{tab:nomenclature} summarizes the nomenclature used in this paper.

\paragraph{AI autophagy.}
Let $M_0$ be a pre-trained LLM and $\mathcal{D}_0$ a dataset of human-authored documents. 
AI autophagy is a process consisting of iterative steps of content generation followed by fine-tuning on the generated data.
In a generation step $j$, the model $M_{j}$ generates a set of documents $\mathcal{D}_{j+1}$. 
In the corresponding fine-tuning step, $M_{j}$ is fine-tuned on $\mathcal{D}_{j+1}$ to produce the next model $M_{j+1}$.
An AI autophagy process of $T$ iterations consists of the sequence $\{M_{j}, \mathcal{D}_{j+1}\}_{j=0}^T$, where each model is fine-tuned on the documents it has generated.
%

\paragraph{Model collapse.}
Given the sequence $\{M_{j}, \mathcal{D}_{j+1}\}_{j=0}^T$, let $\mathcal{P}_{\text{real}}$ denote the true data distribution, and let $\mathcal{P}_{j+1}$ denote the distribution implicitly defined by model $M_j$ through its generations. 
Let $\mathcal{X} = \{x_1, \dots, x_m\}$ be a set of measurable quality dimensions, and let $\Delta_{j+1}(x)$ quantify the deviation of $\mathcal{P}_{j+1}$ from $\mathcal{P}_{\text{real}}$ along aspect $x \in \mathcal{X}$.
We say that model collapse occurs along dimension $x \in \mathcal{X}$ when the deviation $\Delta_{j+1}(x)$ increases over multiple iterations of AI autophagy:
\[
\Delta_j(x) > \Delta_{j-1}(x) \quad \text{for many } j.
\]
In the limit, model collapse may be characterized by unbounded divergence:
\[
\lim_{j \to \infty} \Delta_j(x) = \infty \quad \text{for some } x \in \mathcal{X},
\]
indicating that the model is drifting irreversibly away from the true data distribution. 
The function $\Delta_j(x)$ may take different forms depending on the metric -- for instance, drop in linguistic entropy or rise in repetition rate. 
In existing studies \cite{pappalardo2024survey, xing2025caveats}, model collapse is typically diagnosed when the deviation increases monotonically or persistently across iterations, revealing the cumulative degradation of the LLM's generative capacity.

\medskip
\textbf{Example.} Suppose that $M_0$ generates diverse and semantically rich text ($\mathcal{D}_1$). 
As AI autophagy unfolds, $M_j$ is repeatedly fine-tuned on its own outputs ($\mathcal{D}_{j+1}$) and it begins to produce increasingly repetitive and formulaic sentences (e.g., frequently repeating phrases like "The system is designed to..." or "This method allows for...").
This decline in lexical diversity can be measured by a decrease in linguistic entropy $H$, leading to a growing $\Delta_j(H)$, a signal of model collapse along the linguistic entropy dimension.

\begin{table}[htb!]
\centering
\caption{Summary of the notation used in this paper.}
\label{tab:nomenclature}
\begin{tabular}{ll}
\toprule
\textbf{Symbol} & \textbf{Description} \\
\midrule
$M_0$ & Original pre-trained foundation model \\
$M_j$ & Foundation model at simulation step $j$ \\
\hline
$\mathcal{D}_0 = \{d_1^{\text{\tiny (real)}}, \dots, d_n^{\text{\tiny (real)}}\}$ & Initial dataset of $n$ human-authored documents \\
$P = \{p_1, \dots, p_m\}$ & Fixed set of prompts \\
$\mathcal{D}_{j+1} = \{d_1^{\text{\tiny ({j+1})}}, \dots, d_m^{\text{\tiny ({j+1})}}\}$ & Dataset of prompt-augmented generated documents at step $j$ \\
$\mathcal{D}_{\tiny\text{pool}}^{(j+1)} = \mathcal{D}_0 \cup \cdots \cup \mathcal{D}_{j+1}$
 & Pool of initial dataset and prompt-augmented \\
 &  documents accumulated up to step $j$ \\ 
$\mathcal{T}_{j+1} \sim \mathcal{D}_{\text{pool}}^{(j+1)}$
 & Training set sampled from $\mathcal{D}_{\text{pool}}^{(j+1)}$ to fine-tune $M_j$ into $M_{j+1}$ \\
$s_i^{(j+1)}$ & Continuation generated by $M_j$ for prompt $p_i$ \\
$d_i^{\text{\tiny ({j+1})}} = p_i \Vert s_i^{(j+1)}$ & AI-generated document: concatenation of $p_i$ and $s_i^{(j+1)}$\\
\hline
$\mathcal{P}_{\text{real}}$ & True data distribution (human-authored content) \\
$\mathcal{P}_{j+1}$ & Distribution implicitly defined by model $M_j$ (AI-generated content) \\
\hline
$\mathcal{X}$ & Set of quality dimensions (e.g., entropy, semantic coherence) \\
$\Delta_j(x)$ & Deviation of $\mathcal{P}_j$ from $\mathcal{P}_{\text{real}}$ along dimension $x \in \mathcal{X}$ \\
\hline
$d$ & A document, i.e., a sequence of tokens \\
$w_i$ & The $i$-th token in a document \\
$q_w$ & Model-assigned probability of token $w$ \\
\hline
$k$ & Number of tokens in each prompt (default $k=64$) \\
$L$ & Maximum length (in tokens) of generated documents \\
$\rho$ & Ranking function used to select documents for training \\
\hline
$H(d)$ & Linguistic entropy of document $d$\\
$G(q)$ & Gini coefficient of probability vector $q$\\
$C(q)$ & Collapsed prediction indicator\\
$A_{\text{CI}}$ & Commonsensical Inference Accuracy\\
\hline
$S_{M_j}(d)$ & Surprise of document $d$ given model $M_j$ (perplexity) \\
\bottomrule
\end{tabular}
\end{table}

\section{Autophagy Simulation Framework}
\label{sec:framework}
We propose a framework for simulating AI autophagy, inspired by Shumailov et al.~\cite{shumailov2024ai}.
The framework -- schematized in Algorithm \ref{alg:autophagy_pipeline} -- starts from a pre-trained LLM \(M_0\) and a dataset \(\mathcal{D}_0 = \{ d_1^{\text{\tiny (real)}}, \dots, d_n^{\text{\tiny (real)}} \}\) of \(n\) human-authored documents. It then proceeds iteratively: at each simulation step \(j\), the model \(M_j\) generates continuations \(s_i^{(j+1)}\) from a fixed set of human-authored prompts \(P\), and is subsequently fine-tuned on the resulting prompt–continuation pairs.

During an initialization phase (lines 1-3 of Algorithm \ref{alg:autophagy_pipeline}), we sample $m$ documents from $\mathcal{D}_0$ and truncate each one of these human-authored document $d_i^{\text{\tiny (real)}} \in \mathcal{D}_0$ to $k$ tokens. This yields a set of $m$ prompts $P = \{p_1, \dots, p_m\}$ that remain fixed throughout the entire simulation.
Fixing the prompt set across iterations allows us to analyse model collapse dynamics in a controlled setting.

During the simulation phase (lines 4-9 of Algorithm \ref{alg:autophagy_pipeline}), at each simulation step $j$ (for $j=0, \dots, T$), we perform two operations in sequence:
\begin{enumerate}
    \item \textbf{Generation.} For each prompt $p_i \in P$, we use model $M_{j}$ to generate a continuation $s_i^{(j+1)}$ such that the total length of the combined text does not exceed $L$ tokens (i.e., $|p_i| = k$, and $|s_i^{(j+1)}| \leq L -k$). 
    Here, $s_i^{(j+1)}$  denotes the continuation of prompt $p_i$ generated by $M_{j}$. 
    We then define the AI-generated document as the concatenation of the prompt and the generated continuation: $d_i^{\text{\tiny ({j+1})}} = p_i || s_i^{(j+1)}$, where $||$ denotes sequence concatenation. 
    The set of all such documents forms a new dataset at simulation step $j$, denoted as $\mathcal{D}_{j+1} = \{d_1^{\text{\tiny ({j+1})}}, \dots, d_m^{\text{\tiny ({j+1})}}\}$.


    \item \textbf{Fine-tuning.} $M_{j}$ is fine-tuned on a dataset $\mathcal{T}_{j+1}$ to obtain the updated model $M_{j+1}$.
    The composition of $\mathcal{T}_{j+1}$ depends on the fine-tuning scenario. We consider three scenarios:
    \begin{itemize}
    \item \texttt{AI}: $\mathcal{T}_{j+1} = \mathcal{D}_{j+1}$, only AI-generated documents produced by $M_{j+1}$ are used for fine-tuning;
    \item \texttt{human}: $\mathcal{T}_{j+1} \sim \mathcal{D}_{0}$, fine-tuning is performed using randomly sampled human-authored documents only;
    \item \texttt{mixed}: $\mathcal{T}_{j+1} \sim \mathcal{D}_{\text{pool}}^{j+1}$, documents used for fine-tuning are sampled from a pool comprising the initial human-authored dataset and AI-generated documents accumulated up to step $j$.
    \end{itemize}
    
    %
    In the \texttt{mixed} scenario, $\mathcal{T}_{j+1}$ is sampled according to a ranking function $\rho: \mathcal{D}_{\text{pool}}^{j+1} \rightarrow \mathbb{R}$.
    %
    The function $\rho$ may reflect various desirable properties, such as linguistic diversity or alignment with a target distribution. The top-$z$ ranked documents according to $\rho$ are then selected to form $\mathcal{T}_{j+1}$. This flexible design allows for experimentation with different document selection scenarios and their impact on model collapse across simulation steps. 
\end{enumerate}

By simulating AI autophagy for $T$ steps, alternating between document generation and fine-tuning, we obtain a sequence of LLMs along with a corresponding sequence of datasets of AI-generated documents, $\{M_{j}, \mathcal{D}_{j+1}\}_{j=0}^T $, where $\mathcal{D}_1$ is generated by the original model $M_0$.

Unlike prior work that masks human-authored context \cite{dohmatob2024model, drayson2025machine, shumailov2024ai, kovavc2025recursive, xing2025caveats, guo2023curious}, we preserve the real prompt during fine-tuning to simulate standard prompt-conditioned generation. 
While this reduces the effect of recursive self-training and keeps the setup realistic, it may also partially mask model collapse, as models can over-rely on the first $k$ human-authored tokens through autoregressive dependencies. 
To better isolate the role of AI-generated data, we repeat the analysis in Section 7 by varying $k$, which controls the proportion of human-authored and AI-generated text. 
We find that model collapse becomes more pronounced as $k$ decreases, consistently across all metrics.



\begin{algorithm}[htb!]
\caption{AI Autophagy Simulation Framework}
\label{alg:autophagy_pipeline}
  \SetKwInOut{Input}{Input}
  \SetKwInOut{Output}{Output}
  \Input{Model $M_0$; initial dataset $\mathcal{D}_0 = \{d_1^{\text{\tiny (real)}}, \dots, d_n^{\text{\tiny (real)}}\}$; prompt length $k$; number of iterations $T$, document selection strategy $\Phi \in \{\texttt{AI}, \texttt{human},\texttt{mixed}\}$.
  
  \Output{Sequence of models $M_1, \dots, M_T$ and datasets $\mathcal{D}_1, \dots, \mathcal{D}_{T+1}$}}

  \BlankLine
  \tcc{Initialization phase}
  \ForEach{$d_i^{\text{\tiny (real)}} \in \mathcal{D}_0$}{
    $p_i \leftarrow \text{truncate}(d_i^{\text{\tiny (real)}}, k)$
  }
  $P \leftarrow \{p_1, \dots, p_n\}$ \tcp*{Fixed set of prompts}

\BlankLine
  \tcc{Simulation phase}
  \For{$j \leftarrow 0$ \KwTo $T$}{
    \ForEach{$p_i \in P$}{
      $s_i^{(j+1)} \leftarrow \text{generate\_continuation}(M_{j}, p_i, L - k)$ \;
      $d_i^{\text{\tiny {j+1}}} \leftarrow p_i \Vert s_i^{(j+1)}$ \tcp*{Concatenate prompt and continuation}
    }
    $\mathcal{D}_{j+1} \leftarrow \{d_1^{\text{\tiny {j+1}}}, \dots, d_n^{\text{\tiny {j+1}}}\}$

    $\mathcal{D}_{\tiny\text{pool}}^{j+1} \leftarrow \mathcal{D}_0 \cup \cdots \cup \mathcal{D}_{j+1}$ \tcp*{Accumulate datasets}

    \If{$\Phi = \texttt{AI}$}{
        $\mathcal{T}_{j+1} = \mathcal{D}_{j+1}$}
    \ElseIf{$\Phi = \texttt{human}$}{
        $\mathcal{T}_{j+1} \sim \mathcal{D}_{0}$
    }
    \Else
    {          
        $\mathcal{T}_{j+1} \sim \mathcal{D}_{\tiny\text{pool}}^{j+1}$  \tcp*{\texttt{mixed}}
    }

    $M_{j+1} \leftarrow \text{fine\_tune}(M_{j}, \mathcal{T}_{j+1})$ \tcp*{Fine-tuning}
  }
\end{algorithm}

\ \\ 




\section{Measuring model collapse: a standard measure, and three new ones}
\label{sec:measures}

We assess model collapse using both existing and novel metrics, capturing complementary perspectives. 
As an existing metric, we use linguistic entropy to quantify diversity in AI-generated documents. 
In addition, we introduce two model-intrinsic measures based on next-token probability distributions -- the Gini coefficient and the number of collapsed predictions -- to directly characterize changes in the model’s predictive behaviour.
Finally, we introduce a measure that quantifies the capability of the model to provide commonsensical answers.

\paragraph{\bf Linguistic entropy} 
Linguistic entropy offers a simple yet informative measure of the diversity in the model’s output distribution.
A decline in linguistic entropy over simulation steps indicates that the LLM is producing increasingly repetitive or predictable text, which is a symptom of model collapse. 
While other metrics such as self-BLEU \cite{alihosseini-etal-2019-jointly_selfbleu}, distinct-n \cite{liu2022rethinkingrefiningdistinctmetric}, or perplexity \cite{Xu_2025_perplexity} could also be used, entropy captures the underlying theoretical measure of information content, making it a suitable and parsimonious choice for our analysis.

Formally, given a text document $d$, let $W(d)$ denote the set of unique tokens in $d$.
The normalised linguistic entropy of $d$ is defined as:
\begin{equation}\label{eq:entropy}
H(d) = -\frac{\sum_{w \in W(d)} q_w \log(q_w)}{\log |W(d)|}
\end{equation}
\noindent where $q_w$ is the empirical probability of token $w$ in $d$, computed as the frequency of $w$ divided by the total number of terms in $d$.
The normalisation factor $\log |W(d)|$ ensures that entropy values are comparable across documents with different vocabulary sizes.

For each model $M_j$, and for each human-authored prompt, we measure linguistic entropy and then report the mean value across prompts.

\paragraph{\bf Next-token probability distribution.}
This measure reflects the model's confidence in predicting the next token in a sequence. A high concentration of probability mass on a few tokens indicates a high predictability in text generation and thus a collapsed model.
We query a model $M_j$ to predict the next token for each prompt and extract the probability distribution over the top 100 most likely tokens, yielding one distribution per prompt. From these distributions, we compute two metrics:
\begin{enumerate}

\item The \textbf{Gini coefficient} measures the inequality of the next-token probability distribution. 
A high Gini coefficient indicates that the model concentrates most of the probability mass on a small number of tokens. 
Formally, given a probability vector $q = (q_1, q_2, \dots, q_m)$, the Gini coefficient is defined as:
\[
G(q) = \frac{\sum_{i=1}^{m} \sum_{k=1}^{m} |q_i - q_k|}{2m \sum_{i=1}^{m} q_i},
\]
where $m$ is the number of candidate tokens (here, $m=100$).
$G(q) \in [0, 1]$, where $0$ indicates perfect equality (all tokens are equally probable) and higher values indicate an increasing concentration of probabilities on fewer tokens.

\item The \textbf{percentage of collapsed predictions} captures extreme overconfidence and near-deterministic behavior in next-token prediction, where the probability mass becomes highly concentrated on a single dominant token.
Formally, given a probability vector $q=(q_1,\dots,q_m)$, we define a prediction as \emph{collapsed} if any probability exceeds the threshold $\tau = 0.99$.
Thus, we define the collapse indicator 
$C(q)$ as 1 if the prediction is collapsed:
$$
C(q) = \begin{cases} 1 & \text{if } \exists i: q_i > \tau, \\
0 & \text{otherwise}.
\end{cases}
$$
\end{enumerate}

For each model $M_j$, we report the mean Gini coefficient across prompts and the percentage of prompts flagged as collapsed.

\paragraph{\bf Commonsensical Inference Accuracy} 
A key question is whether model collapse also degrades a model’s ability to produce meaningful sentences. 
To investigate this, we evaluate performance on a commonsense natural language inference task, which tests whether an LLM can plausibly complete a given sentence. We use the HellaSwag dataset~\cite{zellers2019hellaswagmachinereallyfinish}, which consists of $70{,}000$ prompts, each paired with four candidate endings $(\bar{a}, a_1, a_2, a_3)$, where only $\bar{a}$ is a coherent and commonsensical continuation and the others are intentionally implausible.

We compute the conditional probability of each candidate continuation given a prompt. The model selects the option with the highest probability as its predicted continuation.
For each prompt $p \in \{p_1, \dots, p_N\}$, let $\hat{a}$ denote the model’s selected answer and $\bar{a}$ the ground-truth (correct) continuation. We define a binary scoring function $I(p)$ as follows:
$$
I(p) =
\begin{cases}
1 & \text{if } \hat{a} = \bar{a}\\
0 & \text{otherwise}.
\end{cases}
$$
The overall accuracy, which we call Commonsense Inference Accuracy ($A_{\text{CI}}$), is then computed as the proportion of prompts for which the model selects the correct continuation:
$$
A_{\text{CI}} = \frac{1}{N}\sum_{i=1}^{N}I(p).
$$
This measure provides a high-level view of the model’s capacity to make plausible and contextually appropriate decisions, beyond mere stylistic diversity.

\section{Experimental settings}
\label{sec:settings}
Table~\ref{tab:exp_settings} summarizes our experimental settings.
We set the prompt length to $k = 64$ tokens and the maximum length of documents to $L = 128$. This means that we use a 50\% split between human-authored prompt and AI-generated continuation. 
In Appendix~\ref{sec:app_A1_percsynt}, we analyze the effect of varying the proportion of synthetic content by setting $k \in \{32, 96\}$ while keeping $L = 128$ fixed.

To ensure statistical robustness, we run each configuration three times and report the mean and standard error of each metric across runs.
We use 1{,}000 randomly selected prompts from an unseen test set (truncated to $k = 64$) from human-authored datasets to analyze the probability distribution of the first generated token conditioned on each prompt, and compare mitigation results across models.

\paragraph{\bf Large Language Models}
We use four open-weight LLMs: Llama2-7B, Llama3-8B, Mistral-7B, and Qwen3-0.6B, where the suffixes 7B, 8B, and 0.6B denotes the number of parameters in billions.
We report results for Llama2-7B in the main text and defer the others to the Appendix. Llama2-7B offers a strong balance between performance and efficiency, enabling fast fine-tuning while maintaining robust generation quality. 
All models are fine-tuned using parameter-efficient fine-tuning (PEFT) with the Unsloth library using default hyperparameters.
Experiments are run on an NVIDIA Quadro RTX 6000 GPU.

\begin{table}[th!]
\centering
\caption{Experimental settings of our study. The suffix in each model name denotes the approximate number of parameters in billions.}
\label{tab:exp_settings}
\begin{tabular}{@{}p{4cm}p{9cm}@{}}
\toprule
\textbf{Component}         & \textbf{Setting} \\ \midrule
\textbf{LLMs} & 
\begin{tabular}[t]{@{}l@{}}
Llama2-7B \\
Llama3-8B \\
Mistral-7B \\
Qwen3-0.6B 
\end{tabular} \\
\hline
\textbf{Datasets}          & 
\begin{tabular}[t]{@{}l@{}}
Wikitext -- Wikipedia article bodies \\
XL-Sum -- English news article bodies \\
SciAbs -- Abstracts from NLP/CL papers \\
\texttt{combined} -- All three datasets combined
\end{tabular} \\
\hline
\textbf{Fine-tuning scenarios} & 
\begin{tabular}[t]{@{}l@{}}
\texttt{AI} -- Only AI-generated documents \\
\texttt{human} -- Only human-authored documents \\
\texttt{mixed} -- Mixture of AI-generated \& human-authored documents
\end{tabular} \\
\hline
\textbf{Initial human-authored dataset size} & $n=15000$ \\
\hline
\textbf{Number of fixed prompts} & $m=1000$ \\
\hline
\textbf{Prompt length}     & $k = 64$ (default); also tested for $k = 32$ and $k = 96$ \\
\hline
\textbf{Document length} & $L = 128$\\
\hline
\textbf{Fine-tuning library} & Unsloth (with default hyperparameters) \\
\hline
\textbf{Training set size} & $n_{doc}=1000$\\
\hline
\textbf{Number of epochs} & $5$\\
\hline
\textbf{Learning rate} & $1e-4$\\
\hline
\textbf{Decoding Strategy} & Top-k sampling (\texttt{do\_sample=True}, \texttt{temperature=0.6}, \texttt{top\_p=0.9}, \texttt{top\_k=50})\\
\hline
\textbf{Hardware}          & NVIDIA Quadro RTX 6000 GPU \\
\bottomrule
\end{tabular}
\end{table}

\paragraph{\bf Human-authored datasets}
We use the three text datasets used in the study by Guo et al.~\cite{guo2023curious}, ensuring consistency with prior work.
These datasets span different domains and linguistic styles, offering a robust testbed for evaluating model collapse:

\begin{itemize}
\item \textbf{Wikitext} is a large-scale corpus comprising over 100 million tokens extracted from verified English Wikipedia articles~\cite{merity2016pointersentinelmixturemodelswikitext}. Following Guo et al.~\cite{guo2023curious}, we use the main body text of each article, as provided by the dataset on HuggingFace.\footnote{\url{https://huggingface.co/datasets/Salesforce/wikitext}} This dataset is particularly suited for studying long-form factual and encyclopedic language.

\item \textbf{XL-Sum} contains 1.35 million annotated news articles from the BBC in multiple languages \cite{hasan-etal-2021-xl}. 
Each entry includes the title, article body, summary, and article URL. 
We focus on the English-language subset and use the article's body text.
\item \textbf{SciAbs} is derived from a BiBTeX bibliography database of papers published in computational linguistics and NLP venues since 1965 \cite{guo2023curious}. 
The dataset comprises over $40{,}000$ papers, utilizing the text from their associated abstracts.
\end{itemize}
We also consider a combined setting in which the three datasets are merged into a single unified dataset, which we refer to as the \emph{combined} dataset.

\paragraph{\bf Fine-tuning scenarios}
We consider three fine-tuning scenarios that differ in the composition of the training set $\mathcal{T}_j$, allowing us to isolate the effects of AI-generated documents and its interaction with human-authored content:
\begin{itemize}
\item \texttt{AI}: the model is fine-tuned exclusively on AI-generated documents;
\item \texttt{human}: the model is fine-tuned only on human-authored documents, serving as a baseline without exposure to AI-generated documents;
\item \texttt{mixed}: the model is fine-tuned on a combination of AI-generated and human-authored documents, capturing more realistic settings where both sources coexist.
\end{itemize}

\paragraph{\bf Simulation horizon}
For all models, datasets, and fine-tuning scenarios, we simulate AI autophagy over $T = 10$ iterative steps, where each step consists of generating new documents and fine-tuning the model on them. 
This fixed horizon, aligned with prior work~\cite{shumailov2024ai}, allows us to track the progressive effects of AI autophagy and observe the onset and evolution of model collapse.

\begin{figure}[t!]
\centering
\includegraphics[width=0.495\columnwidth]{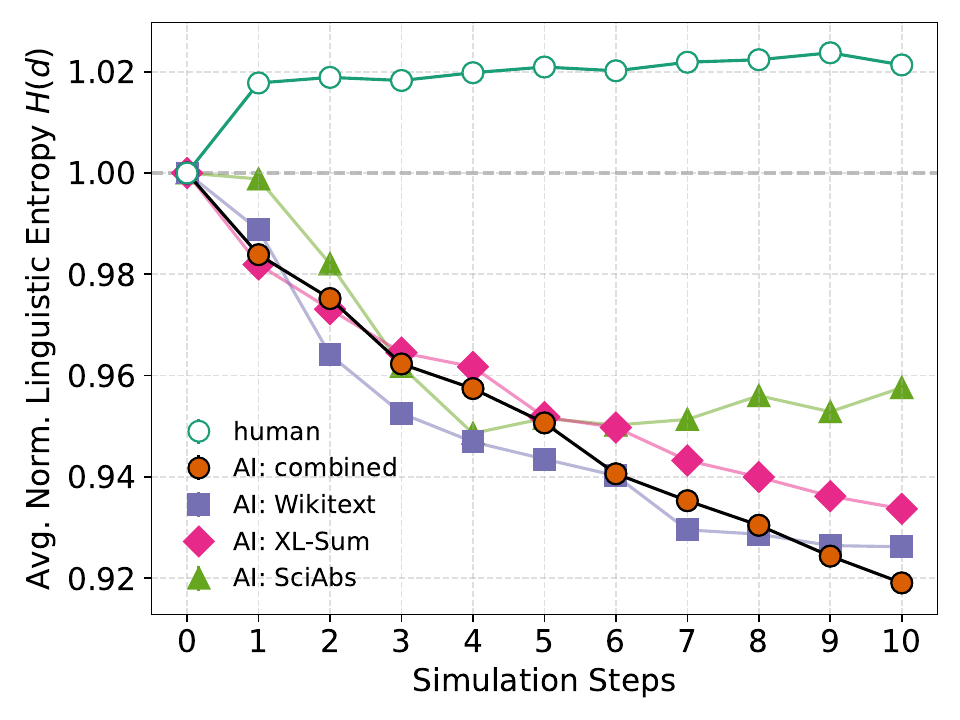}
\caption{\textbf{Effect of AI autophagy on linguistic entropy.}
Normalized linguistic entropy of documents generated by Llama2-7B over 10 simulation steps.
Each curve represents a different fine-tuning scenario: on human-authored documents (green open circles), on AI-generated documents from individual datasets -- Wikitext (squares), XL-Sum (diamonds), and SciAbs (triangles) -- and on a combination of all three (red filled circles).
The results confirm prior findings: linguistic entropy steadily declines as AI autophagy progresses.}
\label{fig:entropy}
\end{figure}

\section{Initial experiments with model collapse – and new observations}
\label{sec:initial_experiments}
Our simulations with Llama2-7B under the \texttt{AI} fine-tuning scenario confirm the findings of prior work~\cite{shumailov2024ai, guo2023curious, pappalardo2024survey}: the average linguistic entropy consistently declines as AI autophagy unfolds, signaling progressive model collapse. 
This decline is evident both when the LLM is fine-tuned on the combined dataset -- with a relative decrease of approximately $7.5\%$ at simulation step $10$ compared to step $0$ -- and when fine-tuned on individual datasets, where the relative decrease ranges from $4.7\%$ on SciAbs to $8.7\%$ on Wikitext (see Figure~\ref{fig:entropy}). 
The reduction in linguistic diversity results from fine-tuning on the AI-generated documents, as no such decline is observed when the LLM is fine-tuned on human-authored documents only (see empty green dots in Figure~\ref{fig:entropy}).
We find similar results for Llama3-8B, Mistral-7B, and Qwen3-0.6B (see Figures \ref{fig:mitigation_allmodels_entropy} and \ref{fig:mitigation_allmodels_hellaswag} in Appendix \ref{sec:app_A4_allmodels}).

\begin{figure}[t!]
    \centering
 \begin{minipage}{0.32\linewidth}
 \subfigure[Step 0, single prompt]
       { \includegraphics[width=\linewidth]{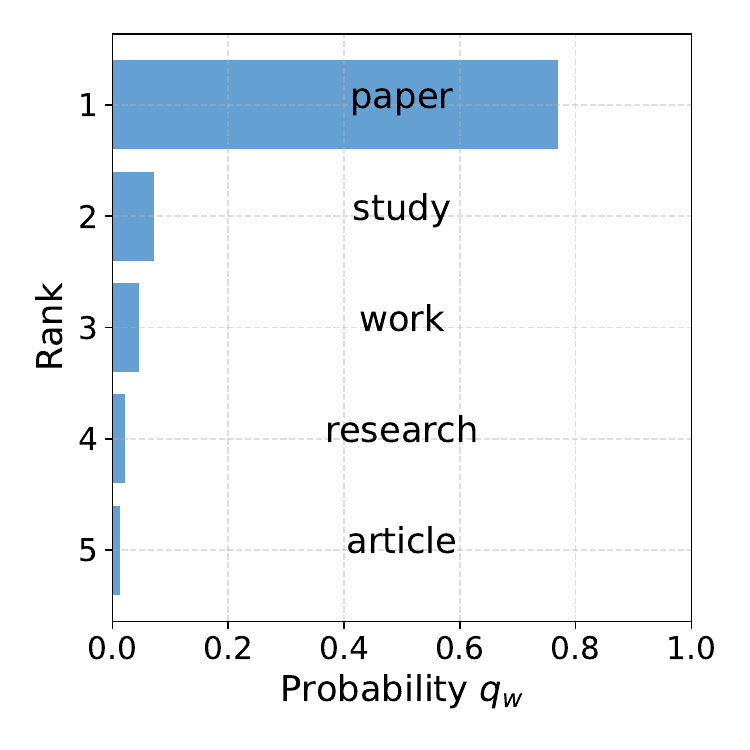}}
    \end{minipage}
    \hfill
    \begin{minipage}{0.32\linewidth}
        \subfigure[Step 5, single prompt]{\includegraphics[width=\linewidth]{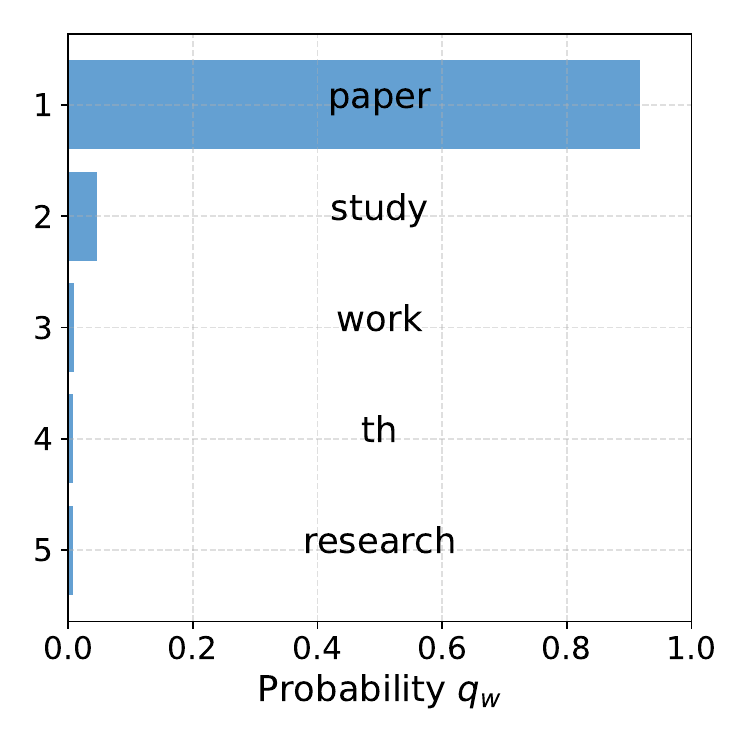}}
    \end{minipage}
    \hfill
    \begin{minipage}{0.32\linewidth}
        \subfigure[Step 10, single prompt]{\includegraphics[width=\linewidth]{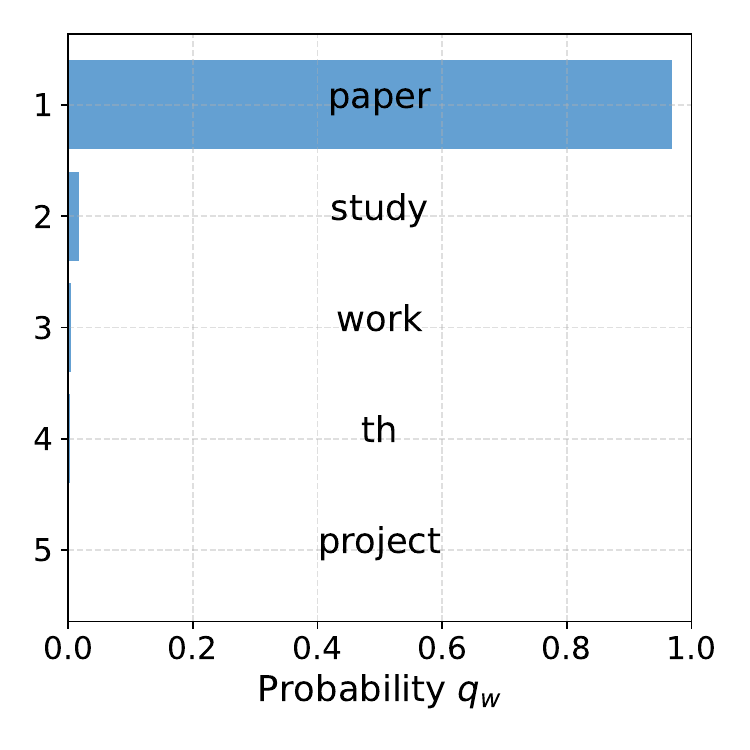}}
    \end{minipage}

    \begin{minipage}{0.32\linewidth}
        \centering
        \subfigure[Step 0, 1000 prompts]{\includegraphics[width=\linewidth]{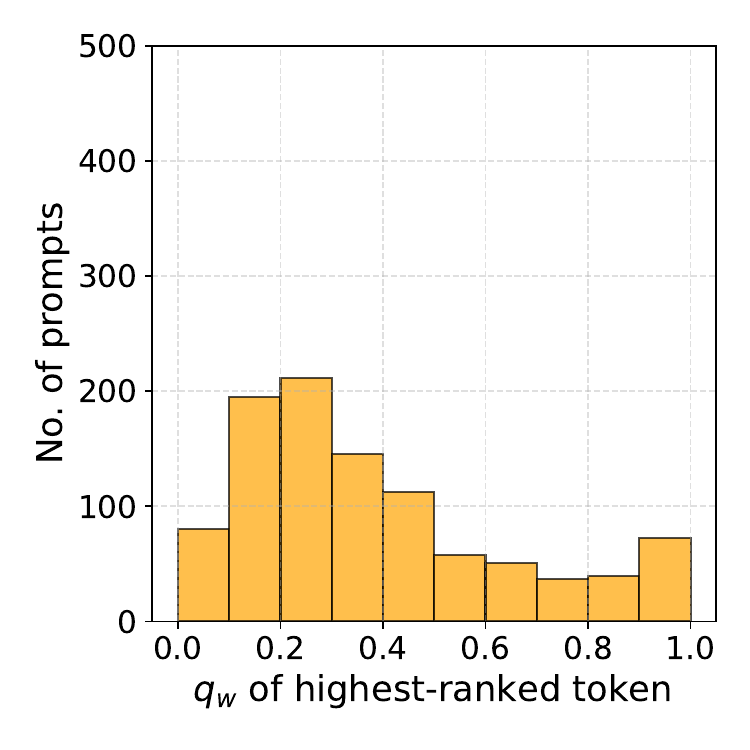}}
    \end{minipage}
    \hfill
    \begin{minipage}{0.32\linewidth}
        \centering
        \subfigure[Step 5, 1000 prompts]{\includegraphics[width=\linewidth]{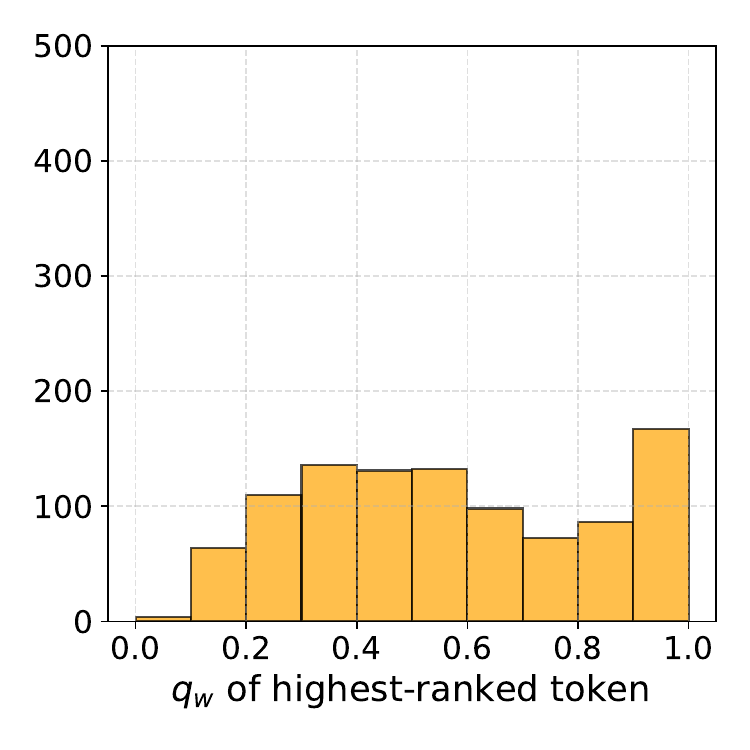}}
    \end{minipage}
    \hfill
    \begin{minipage}{0.32\linewidth}
        \centering
        \subfigure[Step 10, 1000 prompts]    {\includegraphics[width=\linewidth]{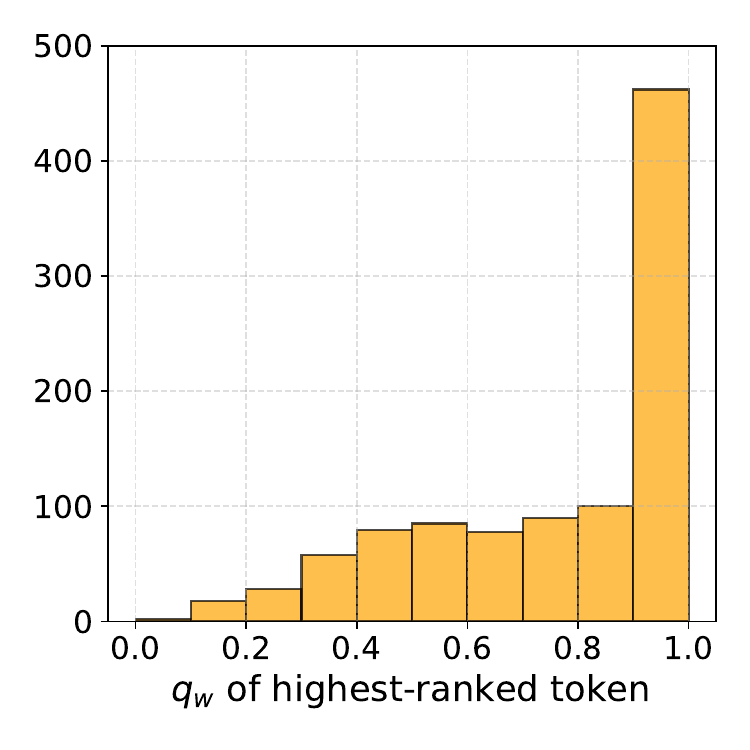}}
    \end{minipage}
\caption{\textbf{Next-token probabilities as AI autophagy progresses.}
(a–c) Next-token probability distributions (x-axis) for the top-5 predicted tokens (y-axis) produced by Llama2-7B at simulation steps 0, 5, and 10.
The example is based on a prompt from the SciAbs dataset:
\emph{“The obstetric Electronic Medical Record (EMR) contains a large amount of medical data and health information. It plays a vital role in improving the quality of the diagnosis assistant service. In this...”}
As the simulation progresses, the distribution becomes increasingly concentrated, with a single token (“paper”) dominating by simulation step 10.
(d–f) Distributions of the highest-ranked token probabilities across 1,000 prompts from the SciAbs dataset, evaluated at simulation steps 0, 5, and 10.
A clear shift toward higher probability values is observed as AI autophagy unfolds.}
\label{fig:nexttoken_probs}
\end{figure}

Model collapse is also evident when analyzing the evolution of next-token probability distributions.
Figure~\ref{fig:nexttoken_probs}a-c illustrates an example with the top-5 predicted tokens and their associated probabilities at simulation steps $0$, $5$, and $10$, using Llama2-7B.
The example refers to text generated in response to the following prompt extracted from the SciAbs dataset:
\begin{quote}
\begin{tcolorbox}[colback=gray!5, colframe=black!30, boxrule=0.5pt, arc=2pt, left=6pt, right=6pt, top=4pt, bottom=4pt]
\textbf{Prompt}: \emph{"The obstetric Electronic Medical Record (EMR) contains a large amount of medical data and health information. It plays a vital role in improving the quality of the diagnosis assistant service. In this..."}
\end{tcolorbox}
\end{quote}

\begin{figure}[!th]
    \centering
    \subfigure[Gini coefficient]
{\includegraphics[width=0.495\columnwidth]{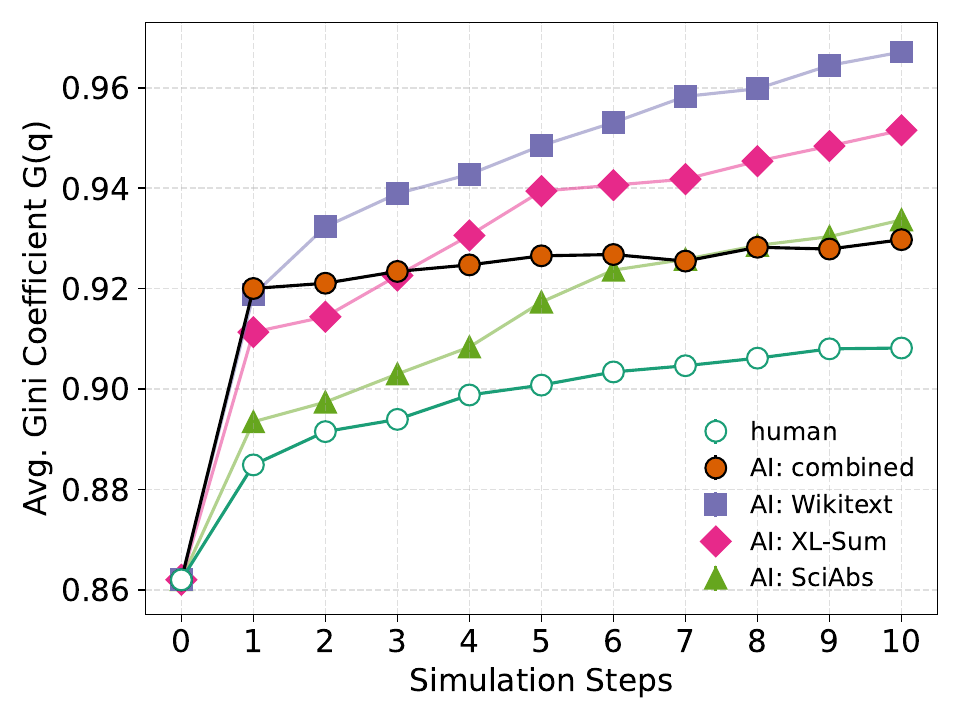}}
    \subfigure[Collapsed predictions]{\includegraphics[width=0.495\columnwidth]{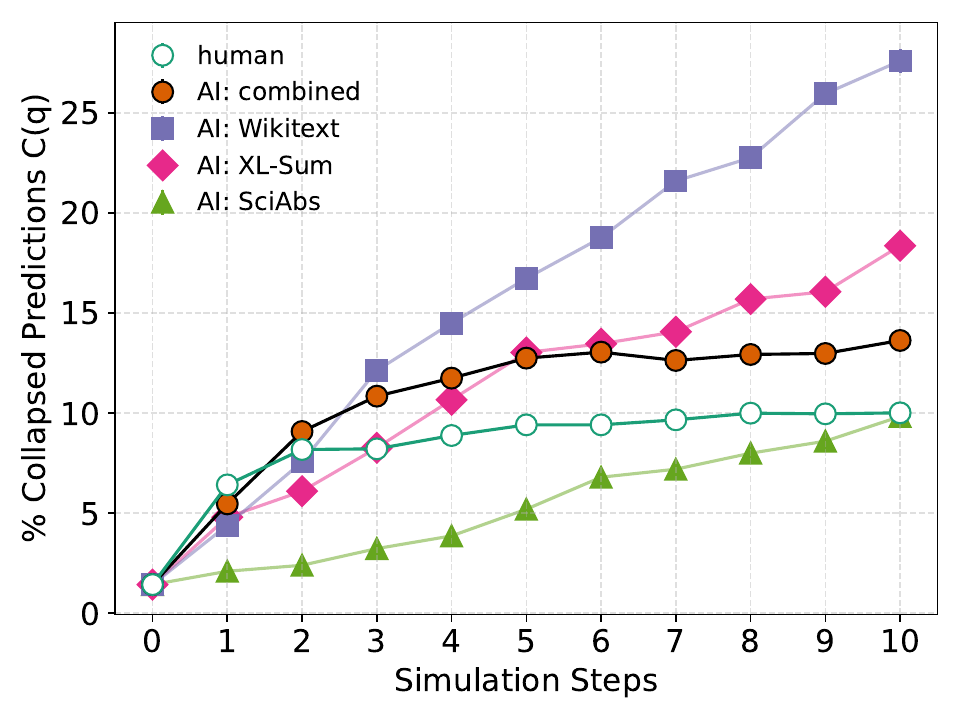}}    \vspace{-1em}
      \caption{\textbf{Effect of AI autophagy on the Gini coefficient and the percentage of collapsed predictions.} (a) Gini coefficient, $G(q)$, of the next-token probability distribution for Llama2-7B across 10 simulation steps.  
      (b) Percentage of collapsed predictions, $C(q)$, across 10 simulation steps.
      Each curve represents a different fine-tuning scenario: on human-authored documents (green open circles), on AI-generated documents from individual datasets -- Wikitext (squares), XL-Sum (diamonds), SciAbs (triangles) -- and on a combination of all three (red filled circles).
      Both $G(q)$ and $C(q)$ increase as AI autophagy unfolds, indicating an increasing concentration of the probability mass on a small subset of tokens.}
    \label{fig:ginicollapse}
\end{figure}

Figure \ref{fig:nexttoken_probs}a-c shows that the distribution becomes increasingly concentrated in simulation steps $5$ and $10$. 
In step $0$, the token ``paper'' has the highest probability but it is still closely followed by other plausible completions. By step $10$, “paper” overwhelmingly dominates the model's next-token probability distribution.

To move beyond a single example, we extend this analysis to a broader set of $1{,}000$ prompts sampled from each human-authored dataset.
We extract these prompts from documents that were never used in any fine-tuning step.
Figure \ref{fig:nexttoken_probs}d–f illustrates how Llama2-7B assigns probabilities to the highest-probability token for the prompts drawn from SciAbs.
As AI autophagy unfolds, we observe a shift toward higher probability values, reflecting a concentration of probability mass on a few tokens.
Figure \ref{fig:other_nexttoken_dists} in Appendix \ref{sec:app_A2_toprankalldataset} shows that the same pattern also applies to Wikitext and XL-Sum. 

This trend is further supported by the evolution of the average Gini coefficient of the next-token probability distributions, $G(q)$, and the percentage of collapse predictions, $C(q)$, over simulation steps.
As AI autophagy unfolds, $G(q)$ increases (see Figure~\ref{fig:ginicollapse}a), indicating an increasing concentration of the probability mass on a small subset of tokens.
A similar pattern is observed for $C(q)$: as the simulation progresses, the percentage of collapsed predictions increases in all fine-tuning scenarios (see Figure \ref{fig:ginicollapse}b). 
Figures \ref{fig:mitigation_allmodels_entropy} and \ref{fig:mitigation_allmodels_hellaswag} in Appendix \ref{sec:app_A4_allmodels} shows that similar results hold for Llama3-8B, Mistral-7B, and Qwen3-0.6B.

\begin{figure}[htb!]
    \centering
\includegraphics[width=0.495\columnwidth]{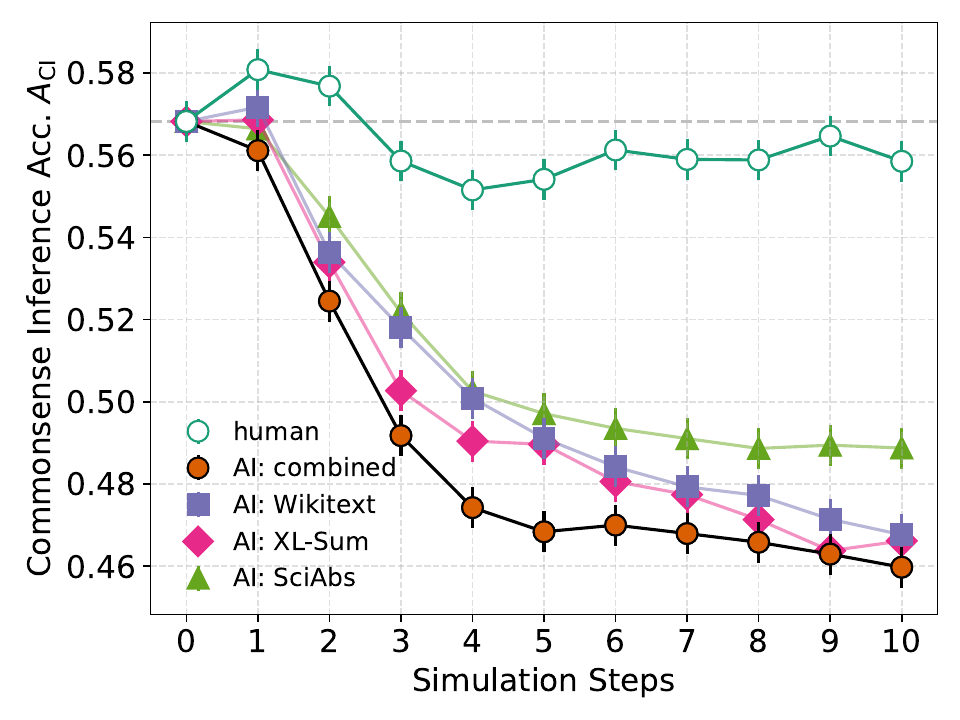}
\caption{\textbf{Effects of AI autophagy on commonsensical inference accuracy.} Commonsensical inference accuracy, $A_{\text{CI}}$, of Llama2-7B across 10 simulations steps.
Each curve represents a different fine-tuning scenario: on human-authored documents (green open circles), on AI-generated documents from individual datasets -- Wikitext (squares), XL-Sum (diamonds), and SciAbs (triangles) -- and on a combination of the three (red filled circles).
As AI autophagy unfolds, the model's predictions increasingly diverge from common sense.}
\label{fig:semantic}
\end{figure}

Model collapse also degrades the model's capacity to generate meaningful sentences.
Figure \ref{fig:semantic} shows that Llama2-7B’s commonsensical inference accuracy, $A_{\text{CI}}$, deteriorates as AI autophagy unfolds.
This degradation is evident especially when the model is trained on the combined dataset -- with a relative decrease of approximately $23\%$ at simulation step $10$
compared to step $0$ -- but also when trained on each dataset individually, with decreases ranging from $1.7\%$ on SciAbs to $16.3\%$ on XL-Sum (see Figure~\ref{fig:semantic}).
In other words, as AI autophagy unfolds, the model generates increasingly implausible continuations of prompts in the HellaSwag dataset.
This result can be attributed to the use of AI-generated documents for fine-tuning, as the decline in commonsensical inference accuracy is markedly less pronounced when the LLM is trained on human-authored documents (see Figure~\ref{fig:semantic}).
Figures \ref{fig:mitigation_allmodels_entropy} and \ref{fig:mitigation_allmodels_hellaswag} in Appendix \ref{sec:app_A4_allmodels} shows that similar results hold for Llama3-8B, Mistral-7B, and Qwen3-0.6B.

Table~\ref{tab:texts} illustrates an example of how the model’s outputs evolve across AI autophagy iterations. 
At simulation step 0, the generated continuations are fluent and meaningful, closely aligned with the prompt. By step 5, signs of model collapse already appear, with increasing redundancy, loss of specificity, and mild incoherence. 
By step 10, model collapse becomes evident: the LLM tends to repeat phrases and produce templated or meaningless sequences.

\begin{table*}[!htb]
\centering
\footnotesize
\def\arraystretch{1.6}
\resizebox{\linewidth}{!}{%
\begin{tabular}{l|p{0.28\linewidth}|p{0.28\linewidth}|p{0.28\linewidth}}
& \large \centering Wikitext & \large \centering XL-Sum & \large \centering SciAbs \tabularnewline
\hline
\multirow{7}{*}{\rotatebox[origin=c]{90}{\bf \normalsize Prompt}} & 
The Church of St George is a medieval Eastern Orthodox church in the city of Kyustendil, which lies in southwestern Bulgaria and is the administrative capital of Kyustendil Province. The church is located in the Kolusha neighbourhood, which was historically separate from the city. The $\dots$
&
At about 19:30 on Friday the fire alarm went off. I
went out of the room and saw other guests milling
about, and no one seemed to be really concerned.
So I went back to my room to get ready to leave
for the airport. I went downstairs to settle the
bill around 20:30 and it was like a scene out of a
movie. 
&
The reliance of deep learning algorithms on large
scale datasets represents a significant challenge
when learning from low resource sign language
datasets. This challenge is compounded when we
consider that, for a model to be effective in the
real world, it must not only learn the variations
of a given sign, but also learn to be invariant to
the person signing. In this paper,  \\
\midrule
\midrule
\multirow{6}{*}{\rotatebox[origin=c]{90}{\bf \normalsize Step 0}} & 
\emph{church
is situated on the eastern side of the city ,
at the foot of the Balkan Mountains . sierp
2011 the church was declared a cultural mon-
ument of national importance . The church is
a single-nave structure with a semi-circular
apse , with a bell tower above the}
&
\emph{I was in the lobby with the receptionist, and the firemen were running down the stairs with their masks on and their hoses. They were shouting, ’Get out, get out!’ I didn’t know what was going on. I thought maybe they were shooting a movie.}
&
\emph{we present a new approach to addressing these challenges, by introducing a novel loss function, which we call the “Mixed Pairwise Loss”, that can be applied to both the training and testing of deep learning models. We present a number of experiments that demonstrate the effectiveness of the proposed method.} \\
\hline
\multirow{6}{*}{\rotatebox[origin=c]{90}{\bf \normalsize Step 5}} & 
\emph{The church is a The church is a The church is a The church is a The church is a The church is a The church is a The church is a</s>} 
&
\emph{The place was filled with smoke and the fire was really big. I was the only guest in the hotel and I was the only one who was able to get out. I was the only one who was able to get out. I was the only one who was able to get out.</s>} 
&
\emph{we propose a novel methodology for learning sign language from a low resource dataset. We propose a novel methodology for learning sign language from a low resource dataset. We propose a novel methodology for learning sign language from a low The reliance of deep learning algorithms on large scale</s>} \\
\hline
\multirow{6}{*}{\rotatebox[origin=c]{90}{\bf \normalsize Step 10}} & 
\hl{\emph{sierp 2020. The church is a The church is a The church is a The church is a The church is a The church is a The church is a The church is a The church is a The church is a</s>}} 
&
\hl{\emph{The place was like a war zone. The place was like a war zone. The place was like a war zone. The place was like a war zone. The place was like a war zone. The place was like a war zone. The place was like a war</s>}} 
&
\hl{\emph{we propose a novel methodology for learning sign language from a low resource dataset. We propose a novel methodology for learning sign language from a low resource dataset. We propose a novel methodology for learning sign language from a low resource dataset. We propose a novel methodology for learning</s>}} \\
\bottomrule
\end{tabular}}
\caption{\textbf{Examples of AI-generated text across AI autophagy steps.} Prompts from the three human-authored datasets are shown at the top of each column. AI-generated continuations by Llama2-7B at steps 0, 5, and 10 are shown in italics, with step 10 additionally highlighted in yellow. The progression illustrates increasing degeneration of generated text over time.}
\label{tab:texts}
\end{table*}

\paragraph{\bf Impact of prompt length}
We assess the impact of the prompt length by varying $k \in \{32, 64, 96\}$, corresponding to 75\%, 50\%, and 25\% of AI-generated tokens (out of a total of $L=128$ tokens). 
We conduct this analysis using Llama2-7B and the combined dataset.

Figure \ref{fig:entropy_semantic_percsynt} in Appendix \ref{sec:app_A1_percsynt} shows the average linguistic entropy, commonsensical inference accuracy, average Gini coefficient and percentage of collapsed predictions across simulation steps for different values of $k$. 
We find that as $k$ decreases, model collapse becomes more pronounced across all the measures. 
In other words, increasing the fraction of AI-generated tokens accelerates model collapse, intensifying the loss of diversity and the overestimation of high-probable tokens.
This result confirms previous findings by \citet{shumailov2024ai}, which show that increasing the proportion of AI-generated tokens exacerbates model collapse.

\section{Learning by surprise: a data  filtering strategy for mitigating model collapse}
\label{sec:mitigation}

An open question is how to mitigate model collapse as AI autophagy unfolds \cite{xing2025caveats}. A strategy that has been proved effective is to incorporate human-authored documents during fine-tuning \cite{xing2025caveats}. However, existing work offers little guidance on how to select such data, typically treating all human-authored texts as equally effective.

We propose a data filtering strategy for selecting documents for fine-tuning that mitigates model collapse, regardless of whether the documents are human-authored or AI-generated.
Our approach is motivated by the observation that model collapse is marked by an increasing inequality in next-token probability distributions, where the LLM becomes increasingly confident in a small set of high-probability tokens.
This imbalance reflects a \emph{loss of surprise}: the LLM is repeatedly fine-tuned on data that closely matches its own expectations, reinforcing its priors and leading to increasingly predictable outputs.
We argue, therefore, that model collapse can be counteracted by intentionally \emph{reintroducing surprise} into the fine-tuning process.

\begin{figure}[!tb]
\centering
\begin{minipage}[t]{0.49\columnwidth}
\centering        \textbf{Wikitext document}\\[0.3em]
\includegraphics[width=\linewidth]{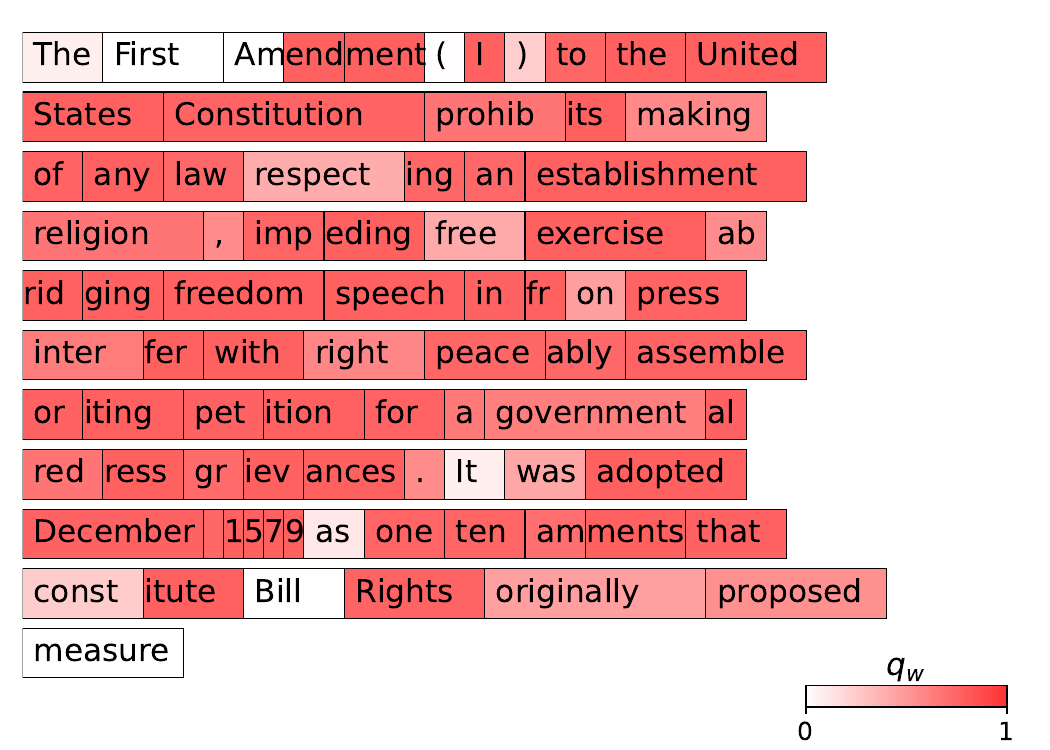}\\[0.4em]
\large Perplexity= 1.6
\end{minipage}
\hfill
\begin{minipage}[t]{0.49\columnwidth}
\centering
\textbf{Sciabs document}\\[0.3em]
\includegraphics[width=\linewidth]{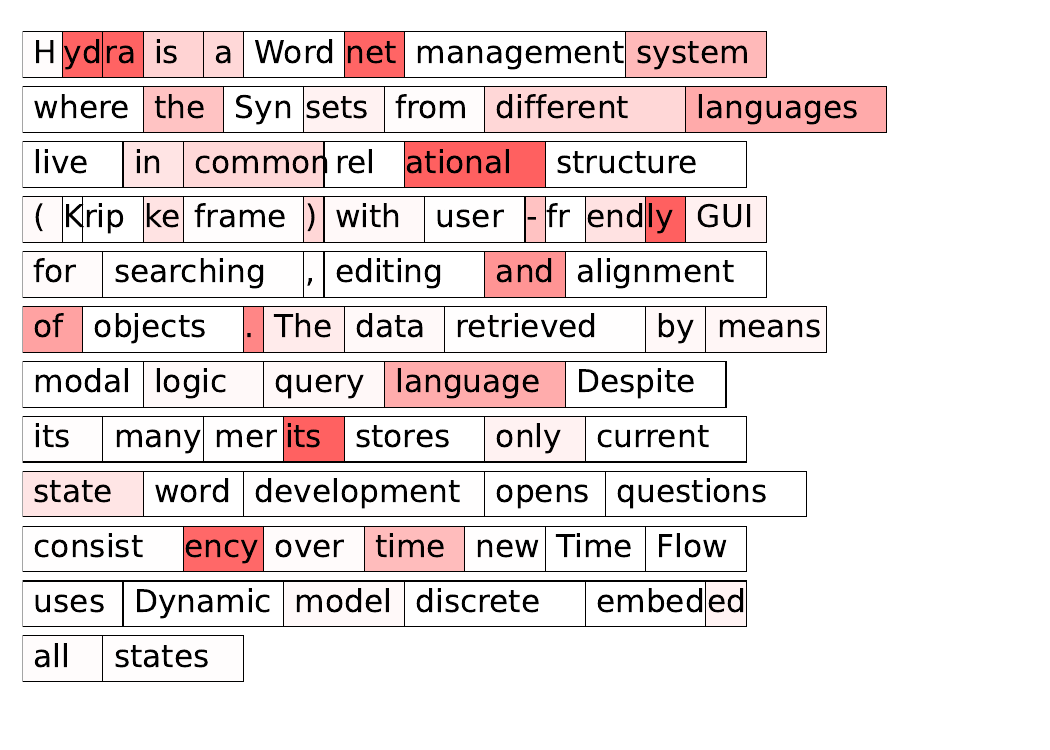}\\[0.4em]
\large Perplexity = 44.5
\end{minipage}
\caption{\textbf{Examples of documents with different perplexity.}
The left document is from the Wikitext dataset, and the right from SciAbs.
Each token is displayed in a box, with background color indicating the probability assigned by the original Llama2-7B model.
The Wikitext document contains consistently higher-probability tokens than the SciAbs document, reflecting a lower perplexity.}   \label{fig:example_surplexity}
\end{figure}

We quantify a model’s surprise using perplexity, where higher values correspond to less expected, and therefore more surprising, content \cite{miaschi-etal-2021-makes}.
Mathematically, given a model $M_j$ and a document $d = (w_1, \dots, w_m)$ composed of $m$ tokens, the perplexity of $d$ given $M_j$ is defined as the exponential of the average negative log-likelihood
of each token given its context:
$$S_{M_j}(d) = \exp \left(-\frac{1}{m} \sum_{i=0}^{m} \log q_i \right)
=\exp(\mathbb{E}[-\log q_i]) = \prod_{i=0}^{m}q_i^{-1/m}$$
\noindent where $q_i = P(w_i \mid w_0, \dots, w_{i-1})$. 
It is important to distinguish between perplexity and linguistic entropy: while both capture aspects of diversity, linguistic entropy is a property of the document alone, whereas perplexity depends on both the document and the model.

To clarify the concept of perplexity, let us consider the two example documents from the Wikitext and SciAbs datasets in Figure~\ref{fig:example_surplexity}.
Each token is shown as a box, with its background color representing the probability assigned to it by the original Llama2-7B.
We observe that the Wikitext document (on the left) contains consistently higher-probability tokens than the SciAbs document (on the right), reflecting a closer alignment with the model’s expectations.
Accordingly, the perplexity score of the Wikitext document given Llama2-7B is considerably lower ($1.6$) than the perplexity score of the SciAbs document ($44.5$).

\begin{figure}[htb!]
\centering
\includegraphics[width=0.495\linewidth]{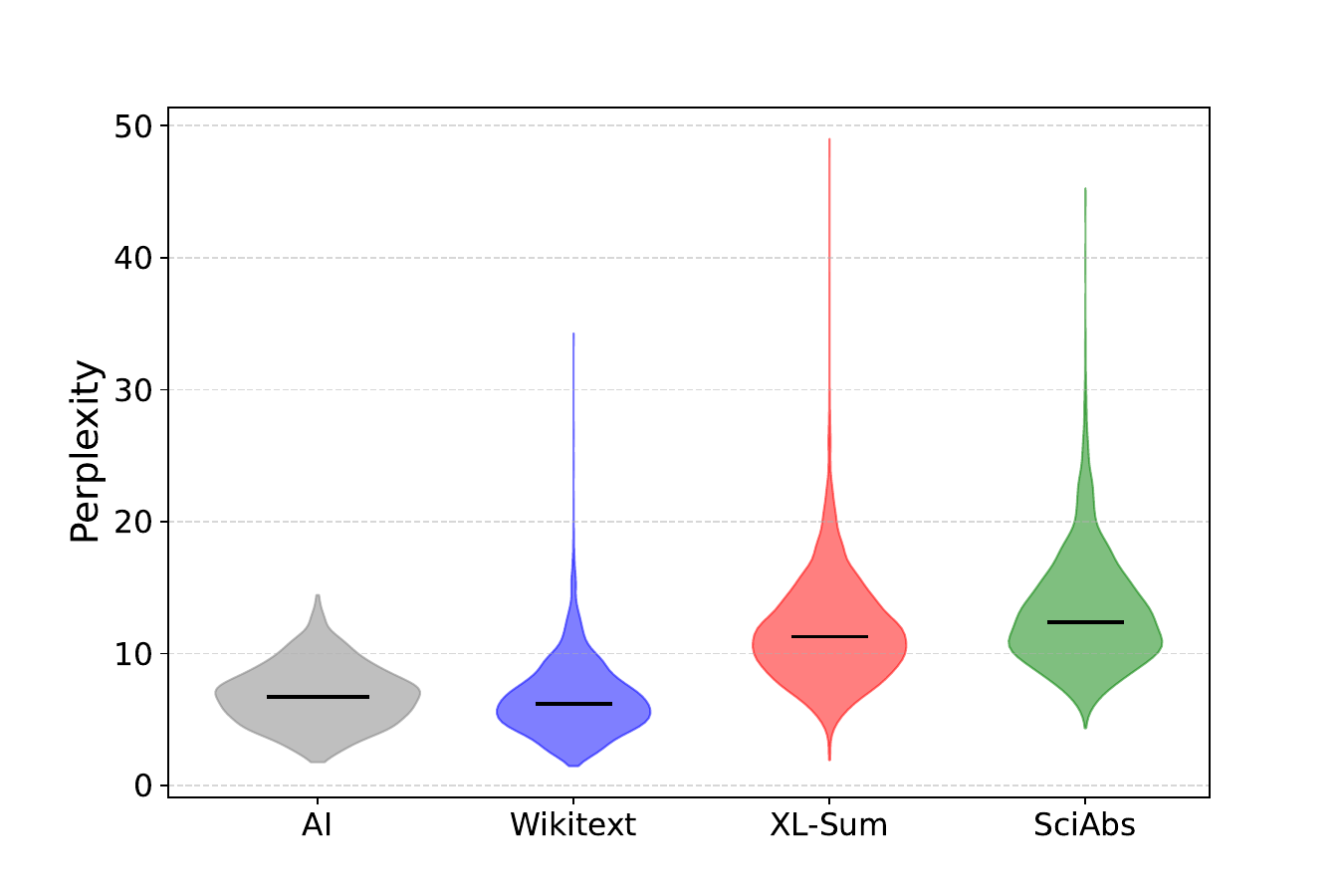}  \caption{\textbf{Distribution of perplexity across the three datasets.}
The perplexity is computed with respect to Llama2-7B for documents in three datasets -- Wikitext (blue),  XL-Sum (red) and SciAbs (green). As a comparison, we provide the distribution of perplexity of documents generated by Llama2-7B itself (light grey).
}
\label{fig:surplexity_on_datasets}
\end{figure}

Beyond this specific example, Wikitext documents are generally less surprising than SciAbs documents.
Figure~\ref{fig:surplexity_on_datasets} shows the distribution of perplexity scores for all documents in the three datasets, computed with respect to Llama2-7B. 
For comparison, we also display the perplexity distribution of AI-generated documents, which typically consists of high-probability tokens and thus lower perplexity.
The Wikitext documents have perplexity scores close to those of the AI-generated documents and substantially lower than those of the XL-Sum and SciAbs datasets.
This suggests that Wikitext documents are inherently less surprising to the model, i.e., they closely reflect Llama2-7B’s internal expectations.

\begin{figure}[t!]
\centering
\subfigure[Linguistic Entropy]
{\includegraphics[width=0.495\columnwidth]{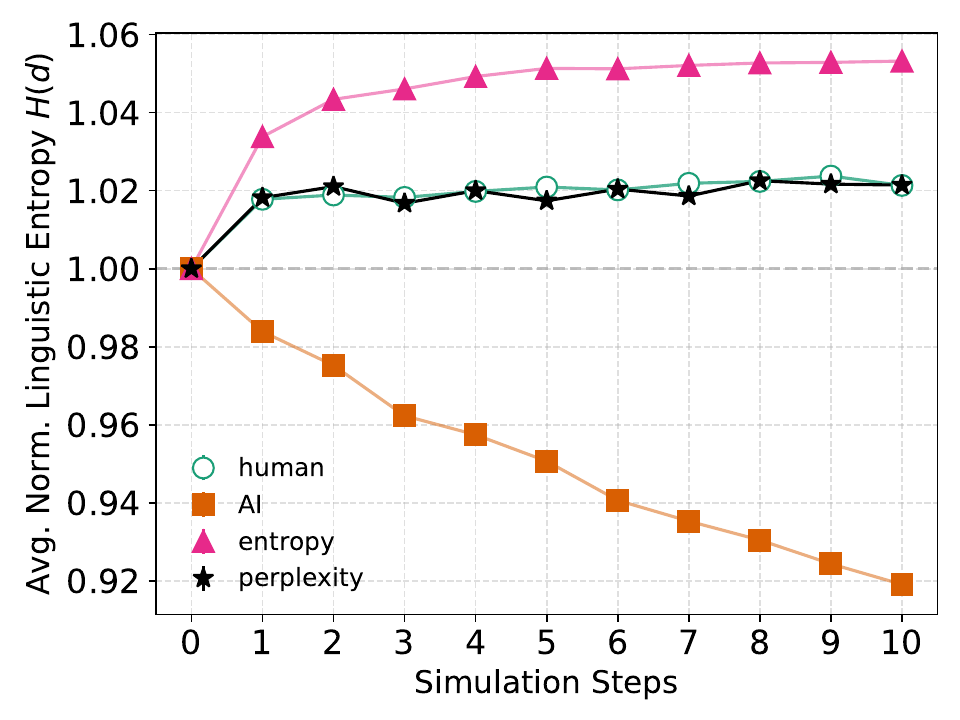}}
    \subfigure[Commonsense Inference Accuracy]{\includegraphics[width=0.495\columnwidth]{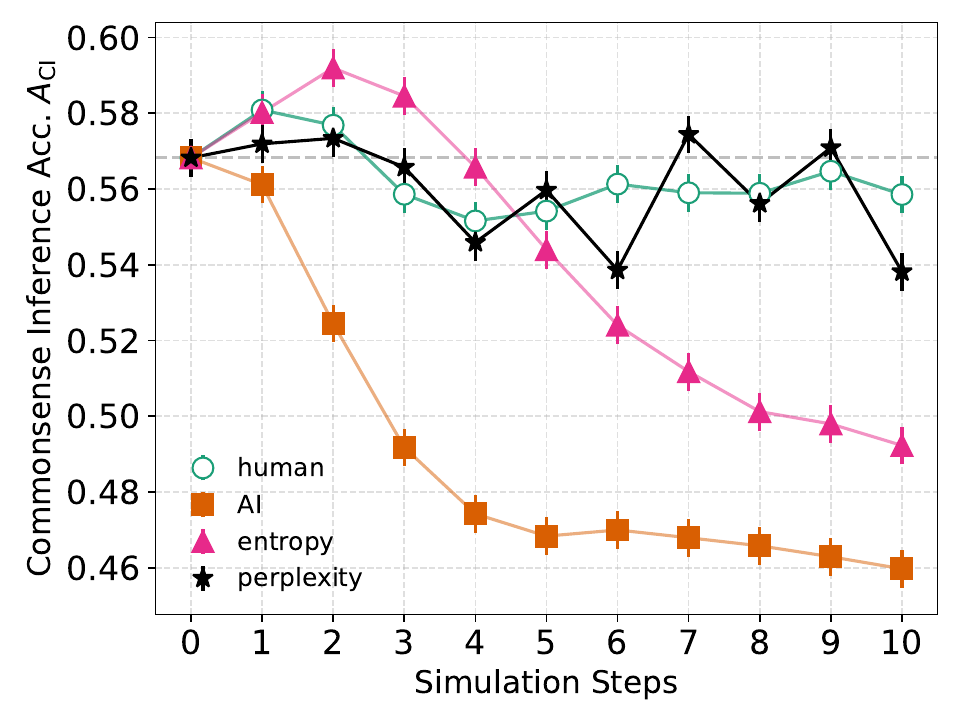}}
     \subfigure[Gini Coefficient]
{\includegraphics[width=0.495\columnwidth]{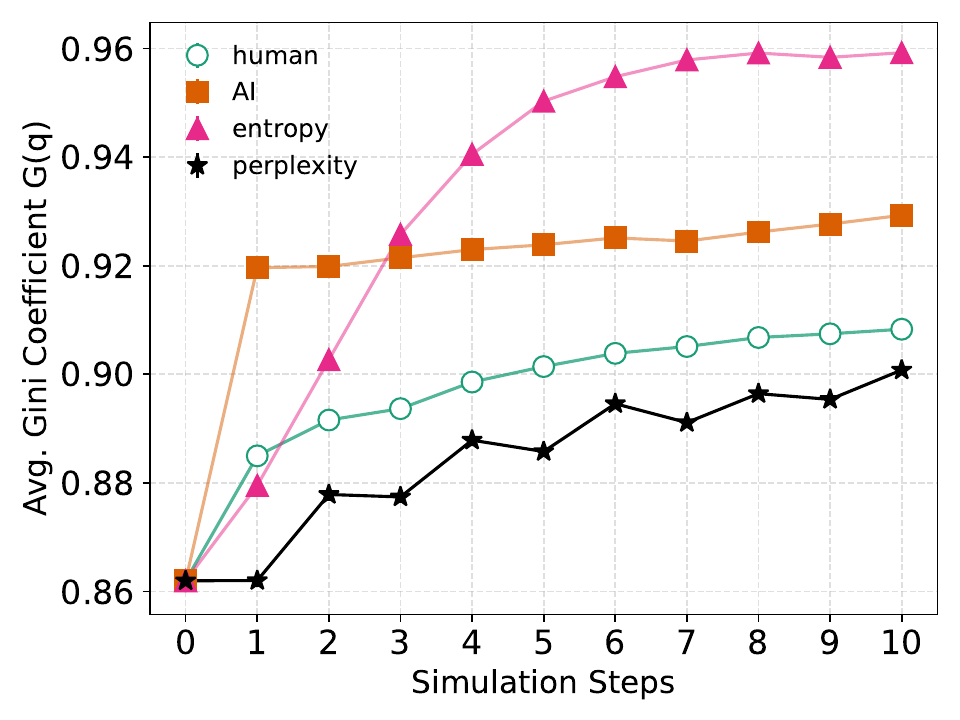}}
    \subfigure[Percentage of collapsed predictions]{\includegraphics[width=0.495\columnwidth]{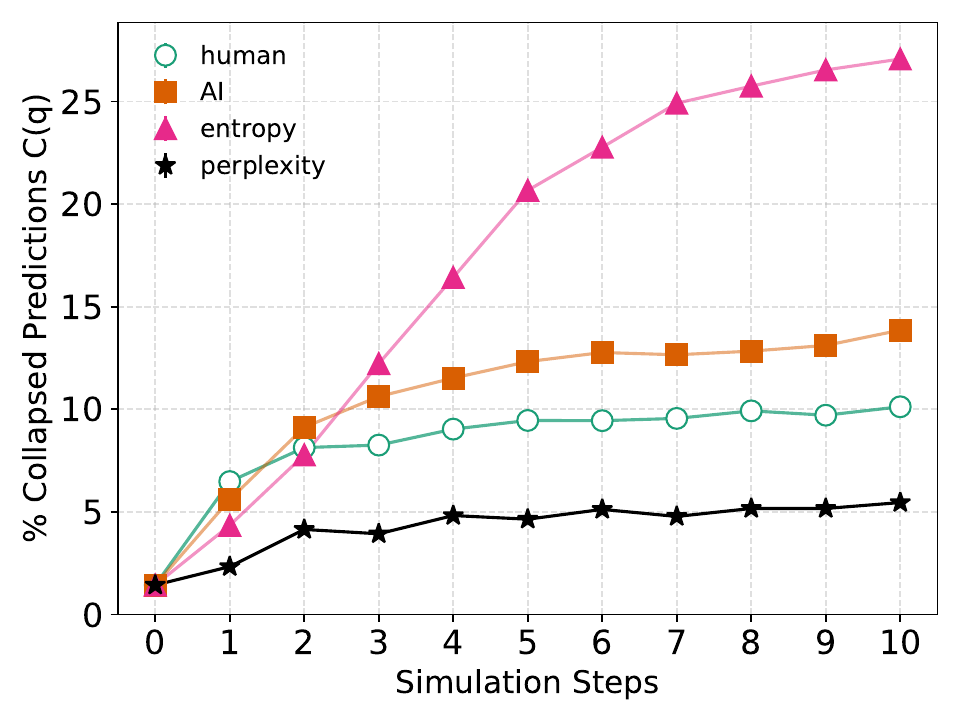}}
\caption{\textbf{Effects of the perplexity-based filtering strategy on model collapse.}
Metrics of model collapse over 10 simulation steps for different mitigation strategies using Llama2-7B.
(a) Average linguistic entropy, (b) commonsense inference accuracy, (c) average Gini coefficient, and (d) percentage of collapsed predictions.
Each curve corresponds to a different mitigation scenario: a random sample of 1,000 human-authored documents (green dots); a random sample of 1,000 AI-generated documents (squares); the 1,000 documents with the highest linguistic entropy, whether human- or AI-generated (triangles); and the surprise-based strategy, selecting the 1,000 documents (whether human-authored or AI-generated) with the highest perplexity (black crosses).}
\label{fig:mitigation_results}
\end{figure}

The Wikitext dataset’s low perplexity is in line with more severe model collapse: as AI autophagy unfolds, models fine-tuned on Wikitext consistently degrade more than models fine-tuned on the other two datasets (see Figures~\ref{fig:entropy} and~\ref{fig:ginicollapse}).
This observation motivates us to hypothesize that model collapse is driven by fine-tuning on documents with low perplexity. 
To address this, we propose a \emph{perplexity-based filtering strategy}: at each simulation step $j$, we evaluate the perplexity of every document of the dataset given model $M_j$ and select the $1{,}000$ documents -- regardless of whether they are human-authored or AI-generated -- with the highest perplexity.
This ensures that fine-tuning targets the most surprising content, independent of document origin.

We compare our perplexity-based strategy against three baselines: \emph{(i)} fine-tuning on a random sample of $1{,}000$ human-authored documents, \emph{(ii)} a random sample of $1{,}000$ AI-generated documents, and \emph{(iii)} the $1{,}000$ documents with the highest linguistic entropy, whether human-authored or AI-generated.
The latter strategy is motivated by prior work, where a decline in linguistic entropy is commonly used to characterise model collapse \cite{shumailov2024ai}. 

As Figure~\ref{fig:mitigation_results} shows, the perplexity-based filtering strategy effectively helps prevent model collapse, achieving an average linguistic entropy comparable to the approach that randomly samples human-authored documents. 
As expected, the entropy-based strategy reaches the highest levels of linguistic entropy, since it is explicitly optimized for that metric.
A similar trend holds for commonsensical inference accuracy: as AI autophagy unfolds, the perplexity-based filtering strategy achieves the highest accuracy, slightly outperforming the baseline where fine-tuning is performed on human-authored documents.
The most striking benefits of the perplexity-based strategy emerge in the average Gini coefficient and the percentage of collapsed predictions: across both metrics, it consistently achieves the lowest values among baseline strategies, indicating a reduced concentration of probability mass on a few tokens.
Similar results hold for Llama3-8B, Mistral-7B, and Qwen3-0.6B (see Figures ~\ref{fig:mitigation_allmodels_entropy}, \ref{fig:mitigation_allmodels_hellaswag}).

\begin{figure}[!htb]
    \centering
    \subfigure[Mitigation Selected Documents]
{\includegraphics[width=0.495\columnwidth]{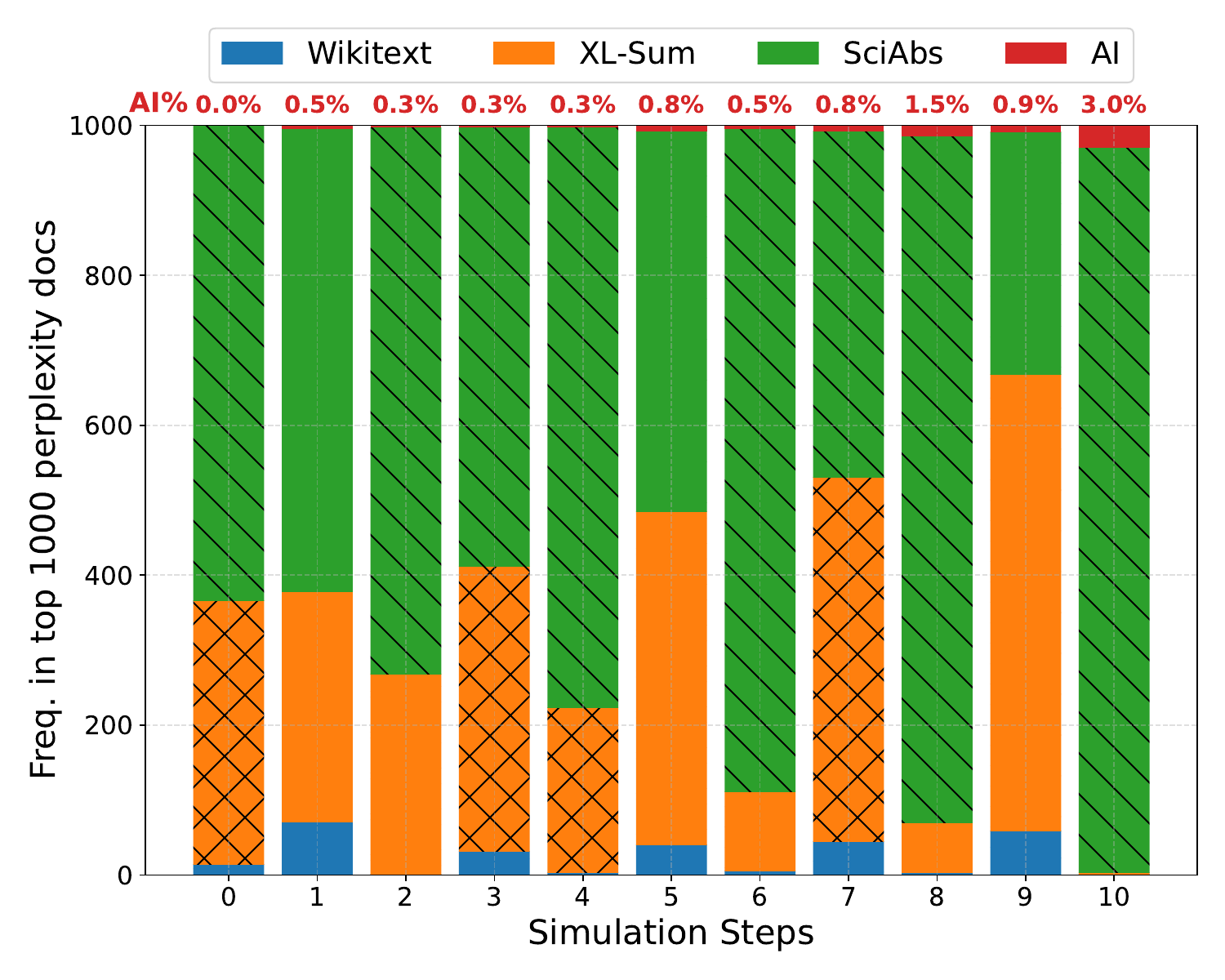}}
    \subfigure[AI Selected Documents]
    {\includegraphics[width=0.495\columnwidth]{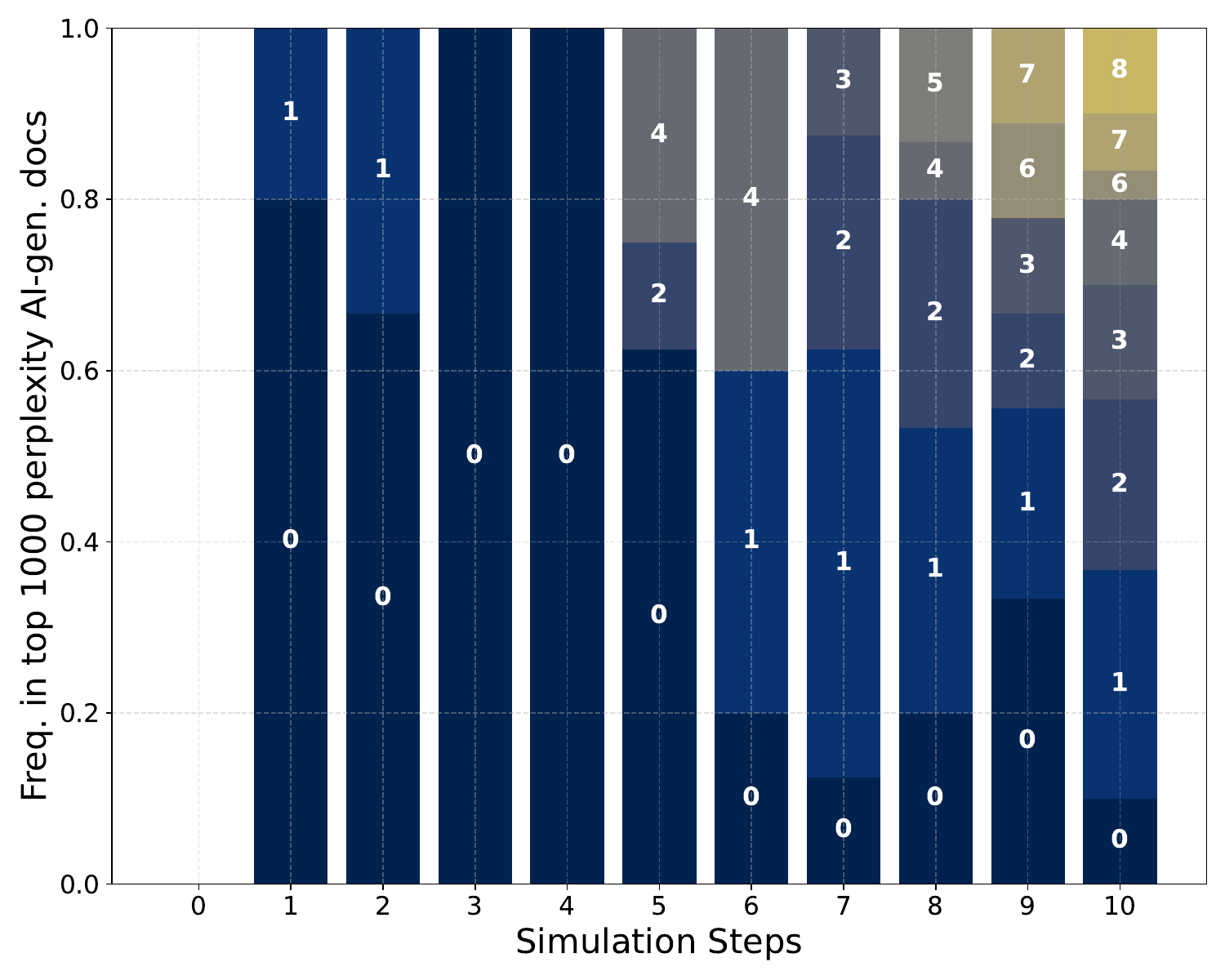}}    
    \vspace{-1em}
      \caption{\textbf{Origin of documents selected by the perplexity-based filtering strategy.} (a) At each simulation step, the bars show the proportion of documents selected for fine-tuning. Wikitext, XL-Sum, and SciAbs denote human-authored documents from the corresponding datasets, while the label ``AI'' denotes AI-generated documents. The percentage of AI-generated documents is reported above each bar. 
      Although the selected documents are predominantly human-authored, the share of AI-generated content increases slightly in later simulation steps. The Wikitext dataset contributes the fewest documents overall, whereas SciAbs and XL-Sum alternate in a complementary, oscillatory pattern across steps.
      (b) For the selected AI-generated documents, the bars at each step show their composition by simulation step. Except for step 1, the selected set of documents does not include those generated in the immediately preceding step. 
}
    \label{fig:mitigation_selected_docs}
\end{figure}

To investigate which documents are selected by the perplexity-based filtering strategy, we analyze the top $1{,}000$ documents (either human-authored or AI-generated) with the highest perplexity at each simulation step $j$. 
Figure~\ref{fig:mitigation_selected_docs} shows that, although nearly all selected documents are human-authored, the share of AI-generated documents increases slightly toward later simulation steps.
Among human-authored sources, Wikitext contributes the fewest documents (less than 100 per step).
While SciAbs documents are generally preferred, the number of selected documents from SciAbs and XL-Sum alternates across simulation steps in a complementary, oscillating pattern.
When model $M_j$ is fine-tuned primarily on XL-Sum, its perplexity on that dataset decreases in the next step, making XL-Sum documents less likely to be selected again. This highlights that the property of causing model collapse is not an intrinsic characteristic of a document, but rather depends on the relationship between the document and the specific model.

Overall, our results show that selecting high-perplexity documents helps mitigate model collapse: since these documents are the least predictable to the model, including them during fine-tuning reintroduces variability and helps rebalance the next-token probability distribution.
Note that our approach represents a clear advantage over existing approaches, as it does not rely on verifying whether or not a document is human-authored -- a task that becomes increasingly challenging as generative AI content proliferates online.

While high-perplexity documents can in principle include highly entropic text -- such as text generated completely at random -- this is not the case in our controlled experiments. 
Indeed, the documents we consider are either human-authored (and thus non-random) or generated by LLMs, and are therefore far from resembling arbitrary noise. 
More generally, the inclusion of truly random documents can be avoided by imposing an upper bound on perplexity.

To complement this discussion, Table~\ref{tab:qualitative} provides qualitative examples of documents selected by the filtering strategy: for step $j=10$, we report the five AI-generated documents with the highest perplexity given $M_{10}$. 
We observe that these documents are predominantly generated in earlier simulation steps (e.g., steps 1 and 2). More generally, at step $j$, selected documents tend to originate from earlier simulation steps, while those generated in the immediately preceding step are less likely to be selected.
Moreover, as Table~\ref{tab:qualitative} shows, high-perplexity documents do not correspond to noise or hallucinated content, but rather to texts that remain informative and well-formed.

%

\medskip

The key distinction of our filtering strategy from prior approaches (e.g., those based on entropy, perplexity, or other scoring functions~\cite{wenzek-etal-2020-ccnet}) lies in its dynamic nature. Rather than computing perplexity once with respect to a fixed model, we recompute it at each step with respect to the evolving model within the AI autophagy process. This yields an adaptive filtering mechanism that tracks model-relative “surprise” as the model distribution changes over time, instead of relying on static notions of data quality.



\begin{table}[h]
\centering
\footnotesize
\begin{tabular}{m{5cm} 
 >{\centering\arraybackslash}m{1cm} 
 m{5cm} 
 >{\centering\arraybackslash}m{1cm} 
 >{\centering\arraybackslash}m{1cm}}\hline
Prompt & Dataset & AI completion & Sim. step & Perplexity ($M_{10}$) \\
\hline We spotted our first guest through a glass wall wandering down a corridor at BBC News HQ, with a flock of assistants around him, back in August 2015. “Shall we grab him? He would be great to kick off the series!” Sir James Dyson had just finished a TV interview in a studio & XL-Sum  &  and we were desperate to start filming the first episode of our new series. But, as we will discover, the British inventor and entrepreneur was not the easiest man to pin down. The first episode of The Dyson Enigma, a three-part series for BBC Two, is about to be broadcast and we have managed to & {\large 1} & {\large 18.63} \\
\hline Soutik BiswasIndia correspondent There, they have been invoking the Constitution and chanting its solemn preamble, which promises justice, equality and fraternity and embodies the basic features of the nation's founding document. The mass readings have revealed a deeper public engagement with the Constitution than commonly thought. So far most believed the Constitution hadn't & XL-Sum  & really inspired popular imagination or created a movement, which is why the Indian Constitution is considered to be "a cold document" by many. The preamble reading initiative has also created a platform for people to ask questions about the Constitution and its provisions. And the answers have been coming in real time. The initiative, started by & {\large 1} & {\large 18.25} \\
\hline We propose DEER (Descriptive Knowledge Graph for Explaining Entity Relationships) - an open and informative form of modeling entity relationships. In DEER, relationships between entities are represented by free-text relation descriptions. For instance, the relationship between entities of machine learning and algorithm can be represented as {``}Machine learning explores algorithms that improve & SciAbs  & the autonomy of decision making.{''} DEER can be viewed as a new type of KG where the node labels are replaced with descriptive texts. In this way, DEER can provide additional explanations to the KG and support human-friendly applications. We propose a new KG completion task to build DEER. Specifically & {\large 2} & {\large 18.25} \\
\hline This work focuses on in-context data augmentation for intent detection. Having found that augmentation via in-context prompting of large pre-trained language models (PLMs) alone does not improve performance, we introduce a novel approach based on PLMs and pointwise V-information (PVI), a metric that can measure the usefulness of a & SciAbs  & contextualized embedding extracted from a sentence. Our PVI-based sampling algorithm uses the salient intent-phrase embeddings extracted from a set of templates that were created using the PLM as a predictor of salience. These embeddings are used to sample new template sentences from the PLM that are informative about the intent category. Our & {\large 7} & {\large 16.63} \\
\hline By Serena KutchinskyNewsbeat online editor The former drug mule appears fragile. The 26-year-old occupies a weird kind of semi-celebrity status. More than 28,000 people who follow her private Instagram account see snaps familiar to many aspiring influencers - a mix of pouting selfies, glam & XL-Sum  & nights out and the odd gym selfie. But behind the filter, the reality is that she is a woman who has spent more than two years in prison for drug trafficking. She is also the subject of a documentary film, Escobar's Prisoner, which follows her journey to the Colombian jungle to meet the drug lord's & {\large 1} & {\large 16.38} \\ 
\hline
\end{tabular}
\caption{\textbf{Qualitative analysis of high-perplexity documents:} at step \( j = 10 \), the table reports the AI-generated documents (excluding human-authored ones) with the highest perplexity with respect to \( M_{10} \), along with the prompt, dataset of origin, generated text, generation step, and perplexity.}\label{tab:qualitative}
\end{table}

\subsection{Relation between perplexity and model collapse}
\label{sec:perplexity-collapse}
To systematically investigate how model collapse varies as a function of perplexity, we construct two datasets: one consisting of $10{,}000$ human-authored documents uniformly sampled from the three datasets (Wikitext, XL-Sum, SciAbs), and one of $10{,}000$ AI-generated documents produced by the original Llama2-7B.
We then sort the documents in each dataset by increasing perplexity given the original Llama2-7B and partition them into $10$ deciles: the first decile contains the 10\% of documents with the lowest perplexity, while the tenth contains those with the highest.
Finally, for each dataset (human-authored and AI-generated), we fine-tune the original Llama2-7B separately on the documents within each decile, yielding 10 models per dataset and 20 models in total.

\begin{figure}[htb!]
\centering
\subfigure[Gini coefficient]
{\includegraphics[width=0.495\columnwidth]{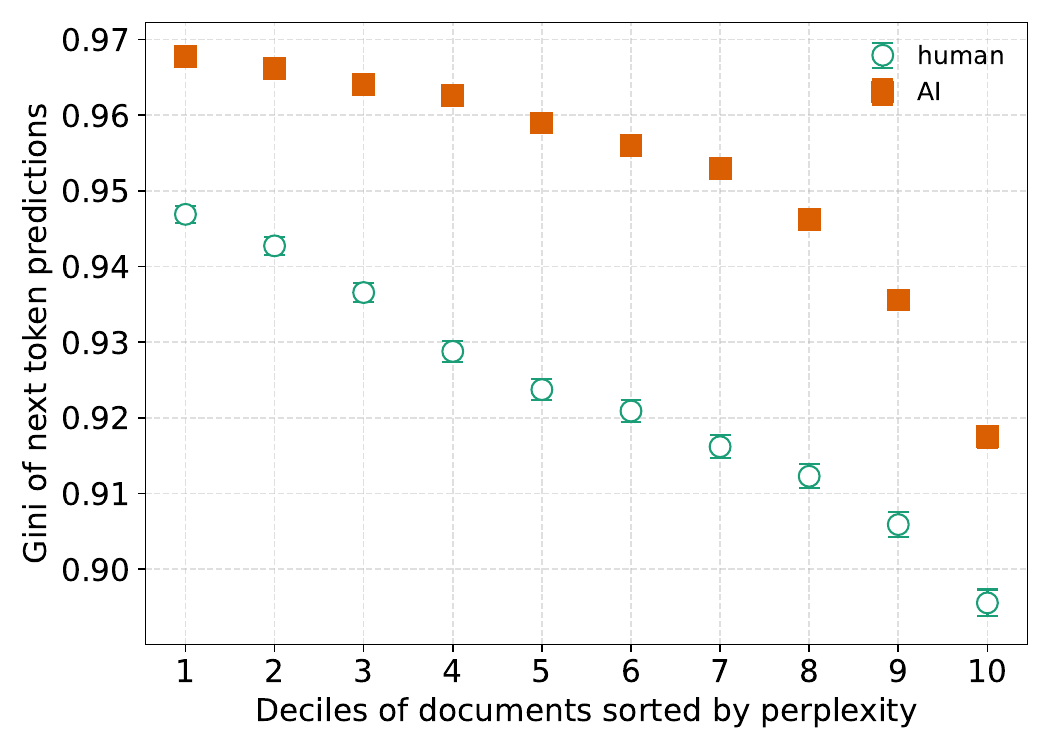}}
    \subfigure[Percentage of collapsed predictions]{\includegraphics[width=0.495\columnwidth]{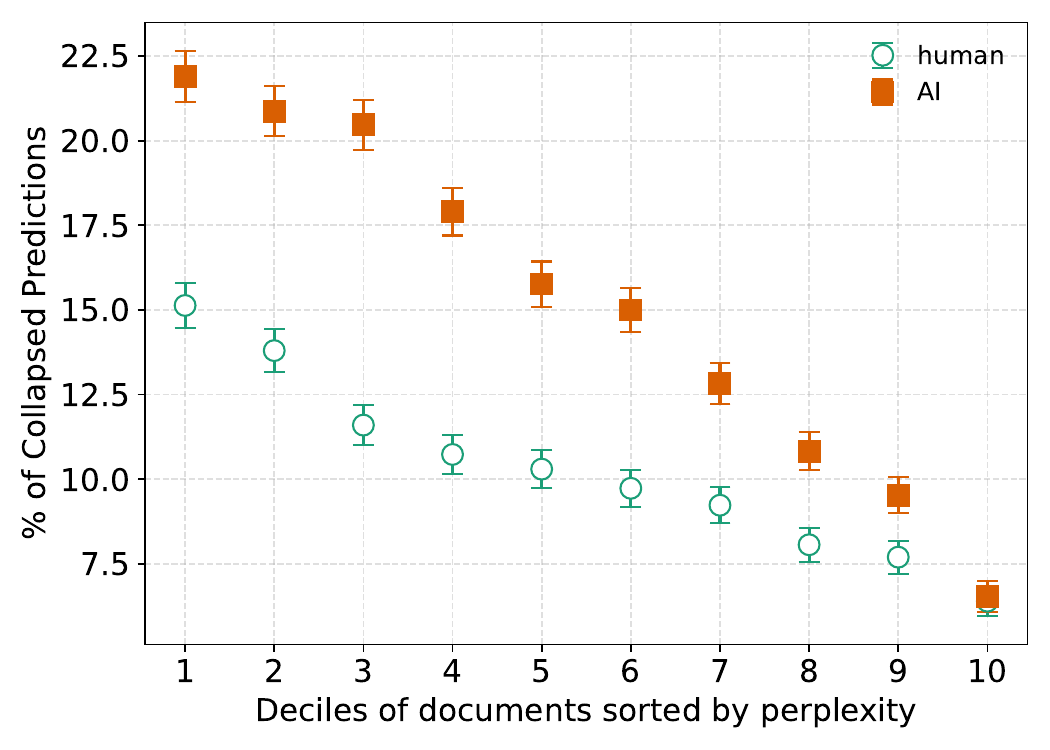}}
   
\caption{\textbf{Relation between perplexity and model collapse} Each point represents a model. The x-axis corresponds to the average perplexity of the i-th decile of documents used for fine-tuning, while the y-axis shows the metric computed on the resulting model. While models fine-tuned on AI-generated documents exhibit greater collapse than those fine-tuned on human-authored documents, in both cases higher document perplexity leads to milder model collapse.}
\label{fig:last_experiment}
\end{figure}

Figure~\ref{fig:last_experiment} shows how the Gini coefficient of the next-token probability distribution (a) and the number of collapsed predictions (b) vary across perplexity deciles, for both the human-authored dataset and the AI-generated dataset.
We find that, while models fine-tuned on AI-generated documents exhibit greater collapse than those fine-tuned on human-authored documents, in both cases higher document perplexity leads to milder model collapse.
In other words, model collapse is more pronounced when fine-tuning on low-perplexity documents. This indicates that the perplexity of the fine-tuning data plays a key role in driving model collapse in LLMs.

\subsection{Computational complexity of the perplexity-based filtering strategy}
\label{sec:compute}
In contrast to existing data filtering strategies, the perplexity-based approach requires computing perplexity for each candidate document prior to fine-tuning. This entails repeated model inference over the candidate pool at each step, which is the dominant source of computational overhead.

To compute the cost of the perplexity-based filtering strategy, we count only the scoring cost, i.e., the forward passes required for filtering, and exclude the fine-tuning cost, since this step is required in any case. In contrast, a static-entropy strategy does not require model inference: the entropy score of each document is computed directly from its token distribution, so its cost comes only from evaluating this document-level statistic and selecting the top-$K$ documents.

Let \(N_0\) denote the initial set of documents, \(K\) the number of new documents generated at each simulation step, \(N_j = N_0 + jK\) the pool of documents at step \(j\), \(\bar{L}_j\) the average token length per document, and \(J\) the total number of simulation steps in the AI autophagy framework. At each simulation step \(j\), we consider the following filtering strategies to select \(K\) documents from the pool \(N_j\):
%

\begin{itemize}
    \item \textbf{Random}: uniformly sample \(K\) documents from the pool. This requires \(O(K)\) time for selection and no scoring cost.
    \item \textbf{Static Entropy}: compute entropy directly for each token. The per-token entropy cost is \(c_{\text{ent}}\), so the total computational cost is:
    \[
    c_{\text{ent}} \sum_{j=0}^{J} N_j \bar{L}_j + O(JK),
    \]
    where \(O(JK)\) is the additional selection cost and
    \(c_{\text{ent}} \sum_{j=0}^{J} N_j \bar{L}_j\) is the main scoring cost.
    \item \textbf{Adaptive Perplexity}: compute perplexity using the current language model \(M_j\). This requires one model forward pass per token, so the per-token scoring cost is \(c_{\text{inf}}\), and the total computational cost is:
    \[
    c_{\text{inf}} \sum_{j=0}^{J} N_j \bar{L}_j + O(JK)
    \]
    Here, \(O(JK)\) is the additional selection cost, and
    \(c_{\text{inf}} \sum_{j=0}^{J} N_j \bar{L}_j\) is the main scoring cost.
\end{itemize}


All three methods include a selection cost, but only perplexity incurs repeated model inference. The key distinction is that static entropy is a text-only computation with per-token cost $c_{\text{ent}}$, whereas perplexity requires model forward passes with per-token cost $c_{\text{inf}} \gg c_{\text{ent}}$.

\paragraph{\bf Practical implications and scalability.}
The repeated rescoring in the perplexity-based filtering strategy makes it more expensive than the static entropy strategy for large \(J\) or \(K\). This may limit the scalability of the perplexity-based strategy on massive datasets or over long training schedules. This overhead could be reduced through several approximations:
\begin{enumerate}
    \item \textbf{Pool subsampling:} At each step, instead of scoring the full pool $N_j = N_0 + jK$, we could score a fixed random subset of size $S = 10K$. 
    This would reduce the per-step cost from $O(N_j)$ to $O(S)$ for scoring, plus $O(K)$ for selection, which is substantially smaller when $S \ll N_j$.
    \item \textbf{Incremental scoring:} We could score only the newly added $K$ documents and reuse the previous step's scores for already evaluated documents. This would reduce the per-step cost to approximately $O(K)$ for scoring, plus the selection cost.
    \item \textbf{Score caching:} For previously scored documents, we could approximate
    \[
    S(M_j, d_i) \approx S(M_{j-1}, d_i) + \Delta S(M_j, d_i),
    \]
    where $\Delta S(M_j, d_i)$ is a lightweight correction estimated from a small subset of documents (e.g., top‑$K$, bottom‑$K$, or a random sample). This avoids recomputing full forward passes for all documents and instead updates scores only where needed. In practice, such caching can reduce redundant computation and keep the overall cost closer to linear in $J$.
\end{enumerate}

Together, these mitigations would make the perplexity-based strategy more feasible in our setting while still capturing model-drift effects, albeit with mild approximations. 
Overall, our perplexity-based filtering strategy is computationally heavier but also more flexible, and it is suitable for moderate-scale, controlled experiments. Scaling to very large $J$ or $K$ would likely require further refinements, such as less frequent re-scoring, progressive sub-pool enlargement, or model distillation, to keep computational costs manageable.

\section{Discussion and Future Works}
\label{sec:conclusions}
In this paper, we addressed four gaps in the understanding of model collapse in LLMs: \emph{(i)} how to characterize a collapsed model directly from its next-token probability distributions; \emph{(ii)} whether AI autophagy causes a loss in the model's ability to generate meaningful and commonsensical outputs; \emph{(iii)} which properties of documents used for fine-tuning contribute to  model collapse; and \emph{(iv)} how to mitigate model collapse without relying on knowing whether data is human-authored or AI-generated. 
We show that a data filtering strategy based on perplexity is effective across various LLMs and datasets, offering a principled, model-centric approach to understanding and mitigating model collapse.

Research on cognitive science shows that surprise drives learning: infants look longer at physically impossible or counterintuitive events, and those events subsequently enhance exploration and learning \cite{margoni2024violation, baillargeon1991object, kidd2012goldilocks}. 
In humans more broadly, Bayesian surprise robustly attracts attention, offering a formal link between surprise and belief updates \cite{itti2009bayesian}, and Friston’s free-energy principle frames perception and learning as minimizing expected surprise over time \cite{friston2010free}. 
Our perplexity-based filtering strategy is an engineering analogue of these findings: by prioritizing high-surprise documents, fine-tuning targets inputs most likely to update the model’s beliefs, rather than reinforcing already-expected outputs, echoing how infants learn most from the unexpected \cite{stahl2015observing}.

We acknowledge that our study has some limitations. 
To ensure computational efficiency and reduce confounding factors, we fixed key fine-tuning parameters such as the number of training documents and the learning rate. These choices are motivated by prior work that has already examined the impact of fine-tuning parameters on model collapse~\cite{herel2024collapse, suresh2024ratemodelcollapserecursive}.
Our analysis is also limited to LLMs and does not extend to generative models for images, audio, video, or multimodal content. While we believe the effectiveness of the perplexity-based filtering strategy may extend to these domains, further validation is needed.

Our simulation setup may introduce biases. 
Since we condition generation on prompts extracted from human-authored text (following the approach of Shumailov et al. \cite{shumailov2024ai}), the model may over-rely on this context, potentially masking effects that would arise in fully synthetic generation loops. 
This could lead us to underestimate exposure bias and model collapse in scenarios where both inputs and outputs are AI-generated.
Finally, as in the broader literature on AI autophagy and model collapse~\cite{pedreschi2024human, pappalardo2024survey}, our findings are based on simulations. Real-world systems are far more complex, with many interacting factors influencing autophagy dynamics. As such, our results should be seen as hypotheses to be tested through carefully controlled experiments on actual generative AI platforms. 

Note that our work suffers from an inherent limitation of working with current open-weight LLMs: model parameters are publicly available, but the exact composition of their training data is not. 
As a result, there is no guarantee that any given evaluation dataset (including commonly used benchmarks) has never been seen during the model's pre-training phase. 
This limitation is not specific to our work and applies to all existing studies on AI autophagy and model collapse. 
A fully controlled held-out evaluation (guaranteeing no overlap with training data) would require models trained on fully transparent and documented datasets. While highly desirable, this is currently not feasible for most state-of-the-art open-weight LLMs, and remains an open challenge for the field.
The effectiveness of perplexity-based filtering may be task-dependent, and prioritizing high-perplexity documents is not guaranteed to be beneficial in all settings. To partially address this, we evaluate our approach not only through distributional metrics (e.g., linguistic entropy and next-token concentration), but also on a downstream task (commonsense inference using HellaSwag). 
We find that perplexity-based filtering does not degrade performance and can even yield slight improvements, suggesting that it preserves useful capabilities in our setting. Nonetheless, this evaluation is limited to a single task, and other tasks may exhibit different trade-offs if high-perplexity data is less aligned with the target distribution. We therefore view our approach as a principled strategy for mitigating collapse, while leaving its broader applicability to future work.

Our perplexity-based filtering strategy is computationally more expensive than simpler alternatives such as static entropy, which scale linearly. 
While this overhead is manageable in controlled experimental settings, it may hinder scalability to larger datasets or longer training regimes. In practice, however, this cost could be mitigated through approximations such as subsampling the pool, incremental scoring, or score caching, which reduce the need for repeated full forward passes.

A limitation of our study is that experiments are conducted at a single scale per model family. While we consider multiple LLMs, each is evaluated at only one size (e.g., 7B for Llama2, 8B for Llama3, 7B for Mistral, and 0.6B for Qwen3). 
The extent to which our findings generalize to larger or smaller models remains an open question, as model size may influence both the dynamics of model collapse and the effectiveness of perplexity-based filtering. Exploring these effects across a broader range of model scales represents an important direction for future work.

Our study opens several directions for future research. We conducted our experiments on three textual domains: general articles (Wikitext), scientific abstracts (SciAbs), and news articles (XL-Sum). 
While these datasets are quite diverse, they are still about human language. It would be valuable to investigate how model collapse unfolds in other contexts such as programming code or mathematical writing, which present unique characteristics. Programming code, for example, has a smaller vocabulary and follows standardized structures and conventions, which may naturally reduce linguistic variety. A similar observation applies to mathematical language, where notation is often fixed and formalization follows well-established norms.
In such settings, AI autophagy might progress even faster, as the baseline content is already low in surprise. 
This is consistent with our findings: in our experiments, the dataset with the lowest perplexity (Wikitext) also exhibited the fastest model collapse. However, low linguistic diversity in domains like programming or mathematics may not be harmful; in fact, it could be beneficial, promoting more consistent formatting and structure. 
Exploring these domain-specific dynamics could provide deeper insights into when collapse is detrimental versus when it may help reinforce useful conventions. We believe this is a promising direction for future work.

From a broader perspective, our results open up intriguing questions about the level of surprise in human-authored content. As our experiments show, different datasets elicit different levels of surprise in LLMs, suggesting that some types of content are inherently more predictable than others. This raises the possibility of characterizing entire content domains -- literary genres, musical lyrics, or scientific texts -- based on how surprising they are to an LLM. 
This line of inquiry could extend beyond text to other human-created media, such as visual art or music, offering a model-based lens through which to explore and compare creative domains. 

Our work offers an innovative contribution to the ongoing debate on AI autophagy and model collapse, and more broadly to the study of human-AI coevolution~\cite{pedreschi2024human}. 
Human-AI coevolution focuses on understanding and modeling feedback loops between humans and AI systems, of which AI autophagy is a prime example. Similar feedback dynamics have been observed in various human-AI ecosystems, such as online retail and urban mapping, where recommender systems can lead to a long-term loss of behavioral diversity among  \cite{pappalardo2024survey, mauro2025urbanimpactaimodeling, cornacchia2022routing, cornacchia2024navigation, fleder2009blockbuster, lee2019recommender}. 
It would be valuable to explore whether this loss of diversity can be formalised as a reduction in surprise and whether the perplexity measure could be adapted to diagnose and mitigate such effects in broader human-AI ecosystems.

\section*{Reproducibility and Code}
All data and models from this study are publicly available through the Hugging Face repository: \url{https://huggingface.co/dgambettaphd}. The simulation framework and analysis presented in this paper are fully reproducible, with code available at
\url{https://github.com/dgambit/LLM_surplexity}.

\section*{Authors’ contributions.} 
DG and GG implemented the code for simulations. DG implemented the code for the data analysis. 
All authors conceptualised the work.
DG and LP designed the figures.
DG made the plots. 
DG, LP, GG, and AK wrote the paper. 
LP and AK directed and supervised
the research. All authors contributed to the scientific discussion, read and approved
the paper.

\section*{Acknowledgments}
Luca Pappalardo has been supported by the Italian Project Fondo Italiano per la Scienza FIS-2024-03129 CAIO.
Dino Pedreschi has been supported by PNRR (Piano Nazionale di Ripresa e
Resilienza) in the context of the research program 20224CZ5X4 PE6 PRIN 2022
“URBAI – Urban Artificial Intelligence” (CUP B53D23012770006), funded by European Union – Next Generation EU.
Gizem Gezici and Fosca Giannotti have been supported by the European Union under ERC-2018-ADG
GA 834756 (XAI), the Partnership Extended PE00000013 - “FAIR - Future Artificial Intelligence Research” - Spoke 1 “Human-centered AI”.

We thank Giuliano Cornacchia, Giovanni Mauro, Gabriele Barlacchi, and Margherita Lalli for the useful discussions.

\bibliographystyle{ACM-Reference-Format}
\bibliography{main}

\newpage
\appendix

\section{Appendix}
\label{sec:app}

\subsection{Different percentage of synthetic tokens}
\label{sec:app_A1_percsynt}

\begin{figure}[htb!]
    \centering
    \subfigure[Linguistic entropy]{\includegraphics[width=0.495\columnwidth]{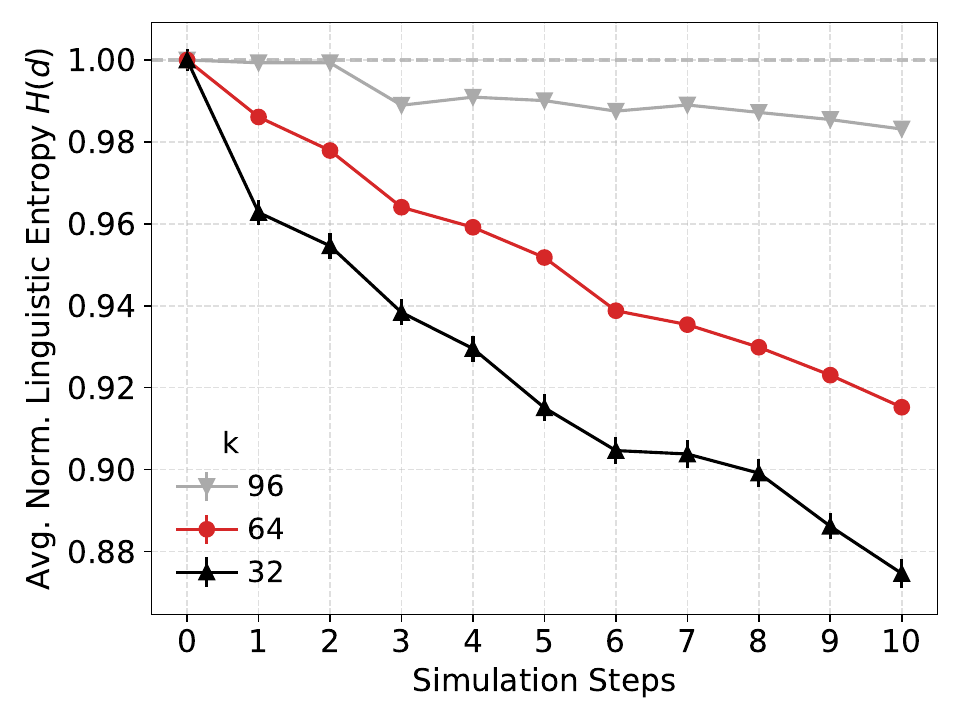}}
    \subfigure[Commonsense inference accuracy]{\includegraphics[width=0.495\columnwidth]{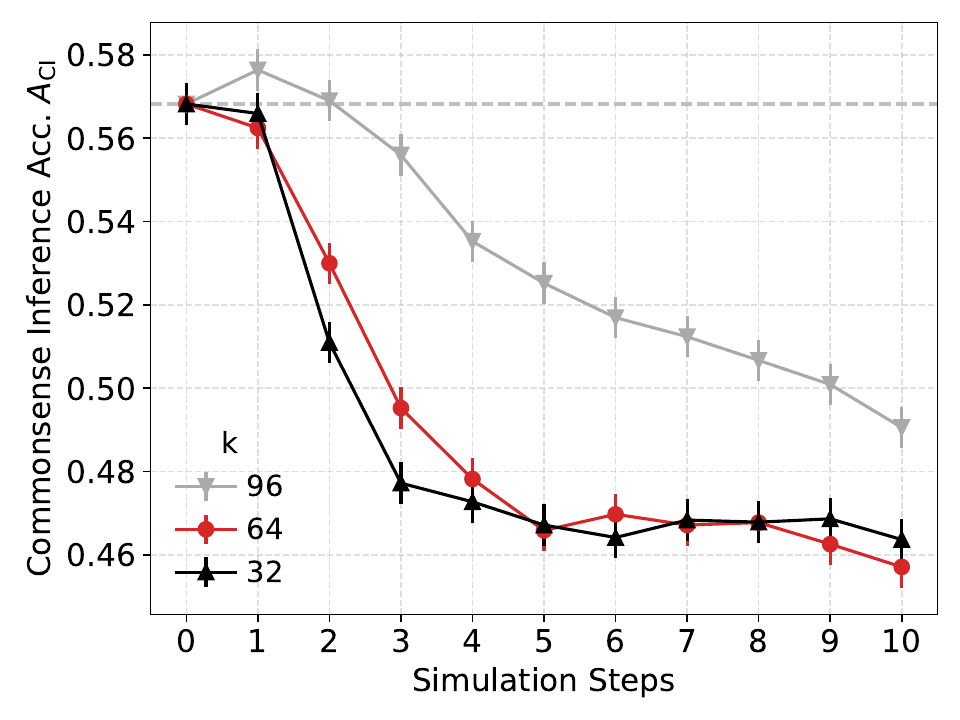}}
    \subfigure[Gini coefficient]
{\includegraphics[width=0.495\columnwidth]{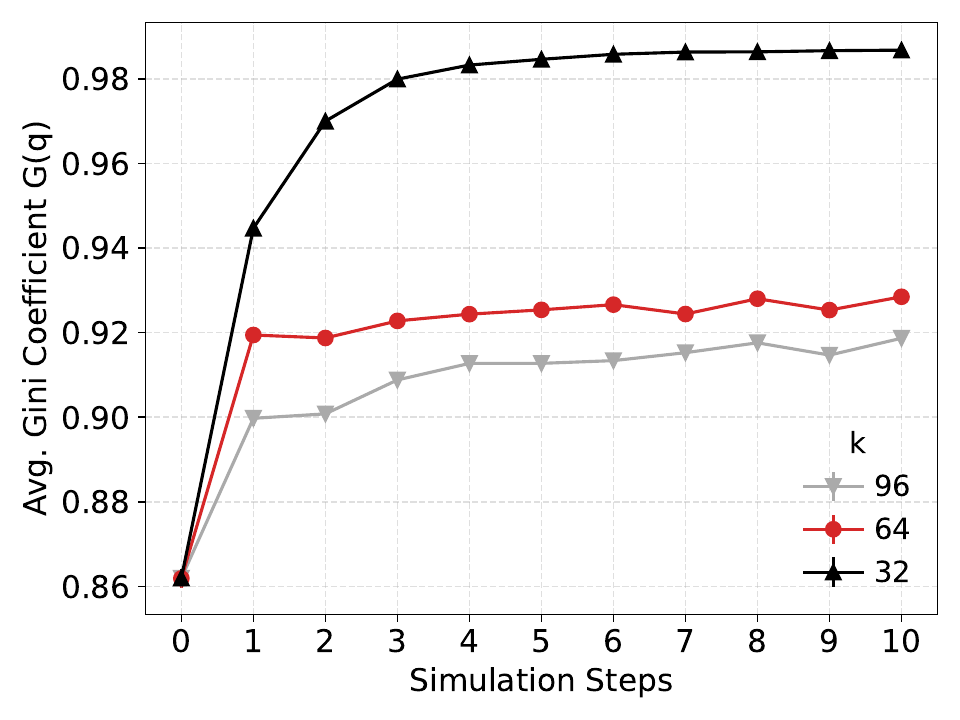}}
    \subfigure[Collapsed predictions]{\includegraphics[width=0.495\columnwidth]{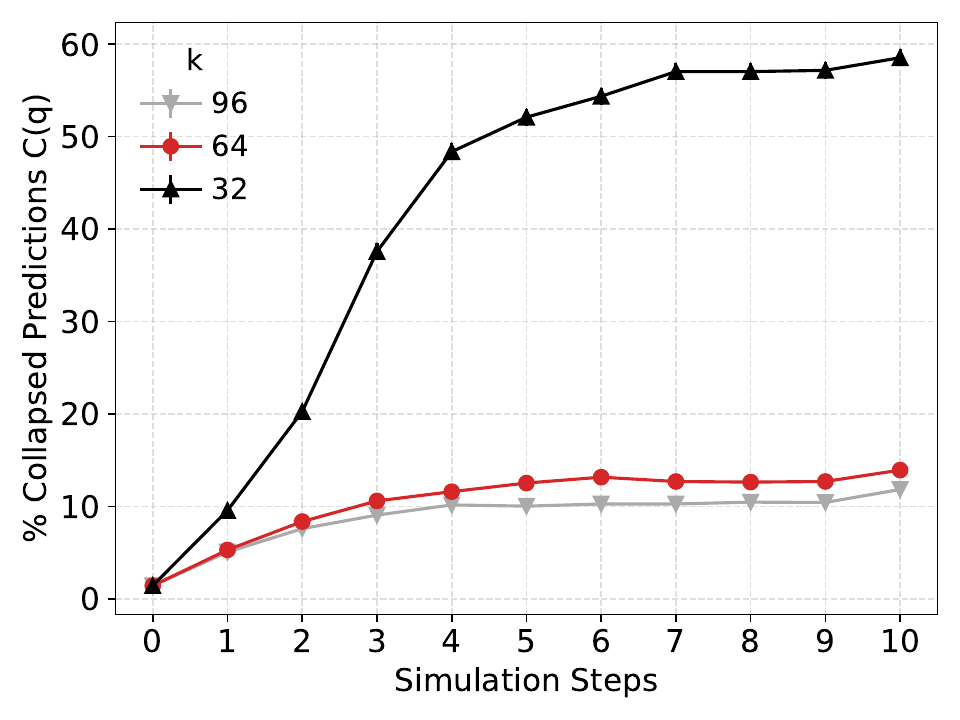}}
\caption{Effects of AI autophagy varying the percentage of AI-generated tokens in generated documents. (a) Normalized linguistic entropy of documents generated by Llama2-7B ($k=32, 46,96$) across 10 simulation steps.
(b) Commonsense inference accuracy of Llama2-7B over the same steps. 
(c) Average Gini coefficient. (d) Percentage of collapsed predictions.
}
\label{fig:entropy_semantic_percsynt}
\end{figure}

In this section, we examine how model collapse varies with the share of AI-generated tokens. 
We fix the total document length at $L=128$ and vary the prompt length $k \in \{32, 96\}$; these correspond to 25\% and 75\% prompt tokens (and thus 75\% and 25\% of AI-generated tokens), respectively. Figure~\ref{fig:entropy_semantic_percsynt} reports linguistic entropy
$H(d)$, commonsensical inference accuracy $A_{\text{CI}}$, average Gini coefficient $G(q)$, and percentage of collapsed predictions $C(q)$ across simulation steps for each $k$. Across all measures, model collapse becomes more pronounced as $k$ decreases, indicating consistent trends among the four measures.

\subsection{Top-ranked next-token probabilities for all datasets}
\label{sec:app_A2_toprankalldataset}

\begin{figure}[htb!]

    \begin{minipage}{0.09\linewidth}
        \centering
\rotatebox{90}{\textbf{Wikitext}}        \end{minipage}
    \begin{minipage}{0.29\linewidth}
        \centering
        \subfigure[]
        {\includegraphics[width=\linewidth]{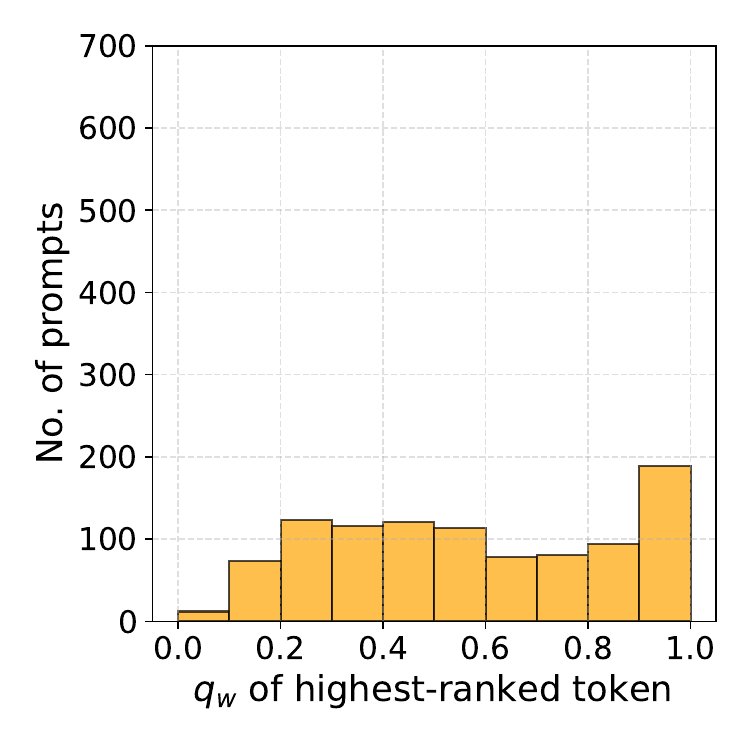}}
    \end{minipage}
    \hfill
    \begin{minipage}{0.29\linewidth}
        \centering
\subfigure[]        
        {\includegraphics[width=\linewidth]{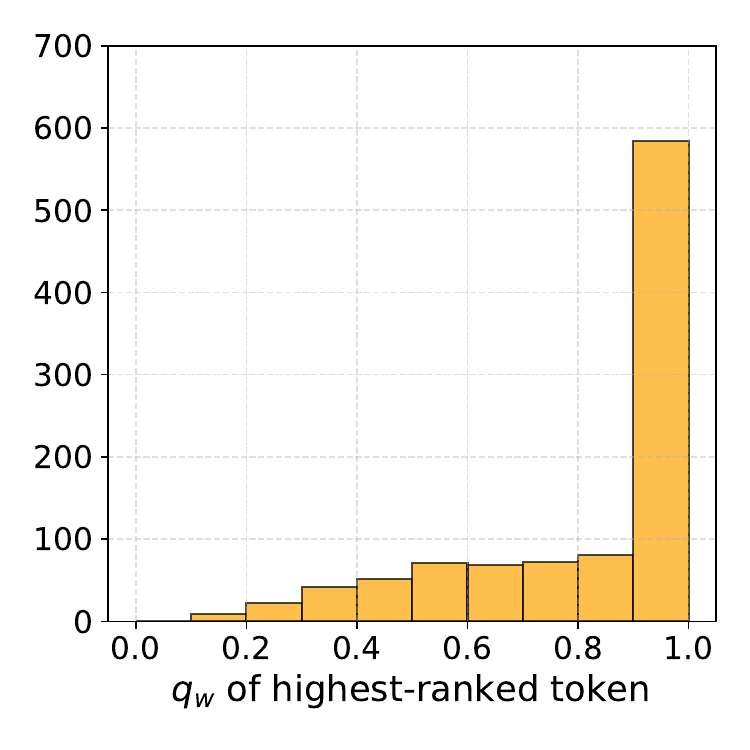}}
    \end{minipage}
    \hfill
    \begin{minipage}{0.29\linewidth}
        \centering
        \subfigure[]
        {\includegraphics[width=\linewidth]{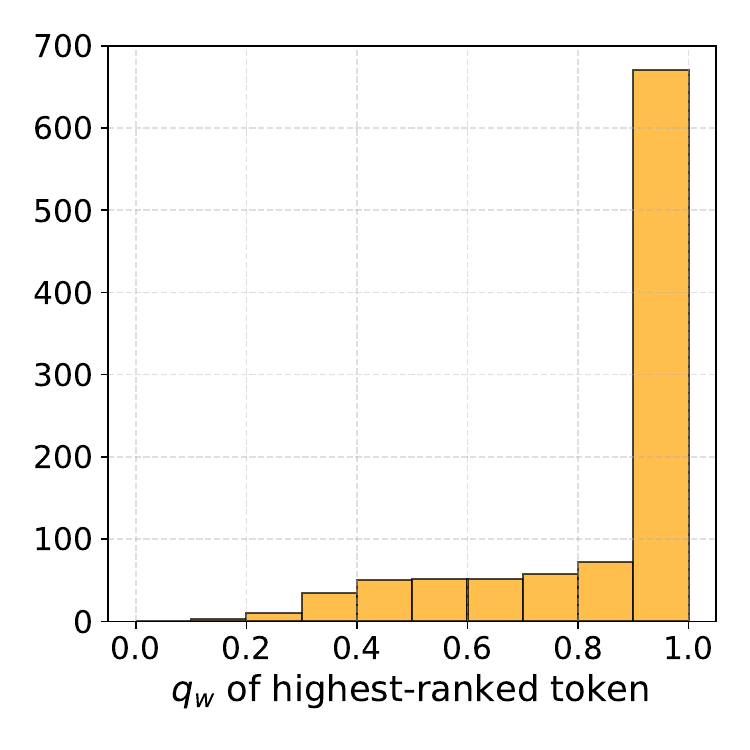}}
    \end{minipage}

    \vspace{0.5em} 

       \begin{minipage}{0.09\linewidth}
        \centering
\rotatebox{90}{\textbf{XL-Sum}}        \end{minipage}
    \begin{minipage}{0.29\linewidth}
        \centering
        \subfigure[]
        {\includegraphics[width=\linewidth]{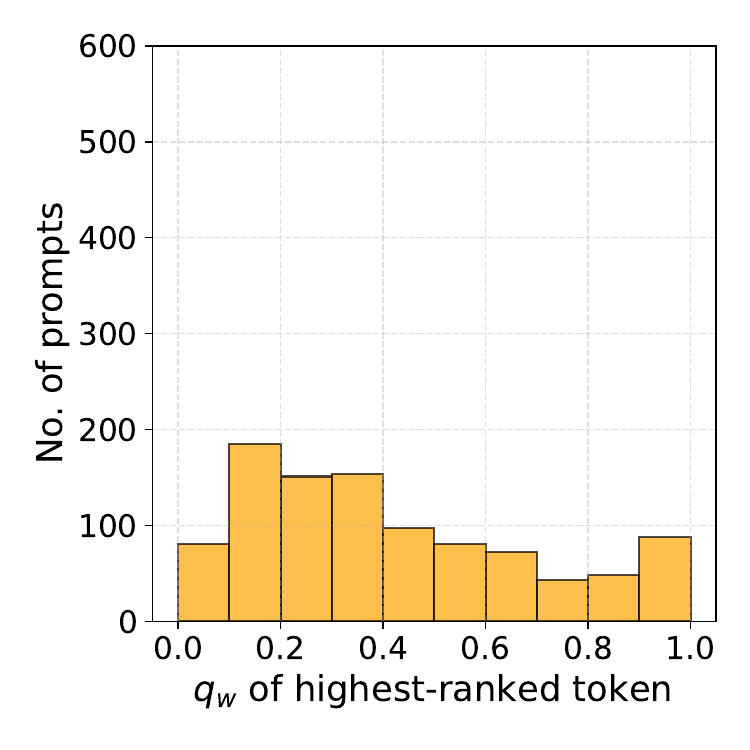}}
    \end{minipage}
    \hfill
    \begin{minipage}{0.29\linewidth}
        \centering
\subfigure[]        
        {\includegraphics[width=\linewidth]{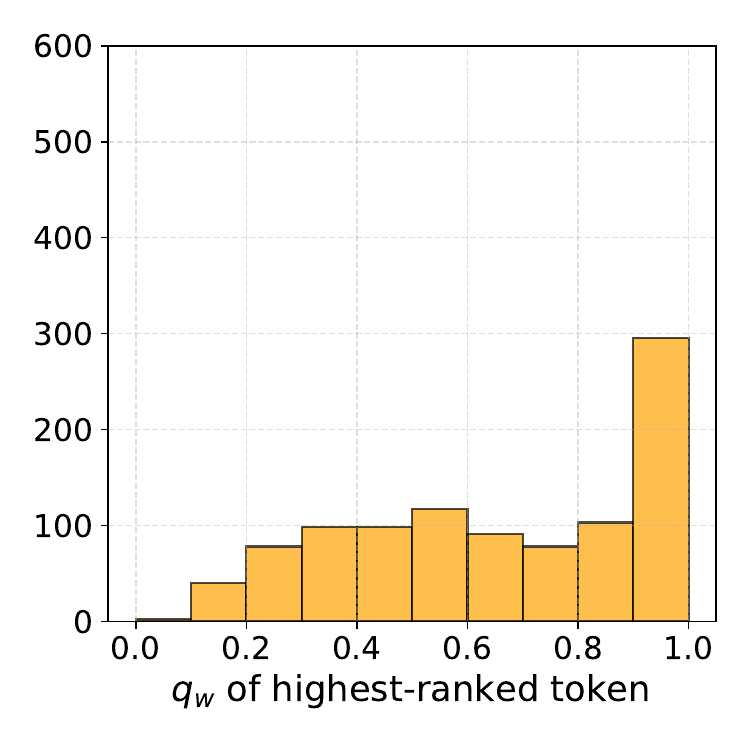}}
    \end{minipage}
    \hfill
    \begin{minipage}{0.29\linewidth}
        \centering
        \subfigure[]
        {\includegraphics[width=\linewidth]{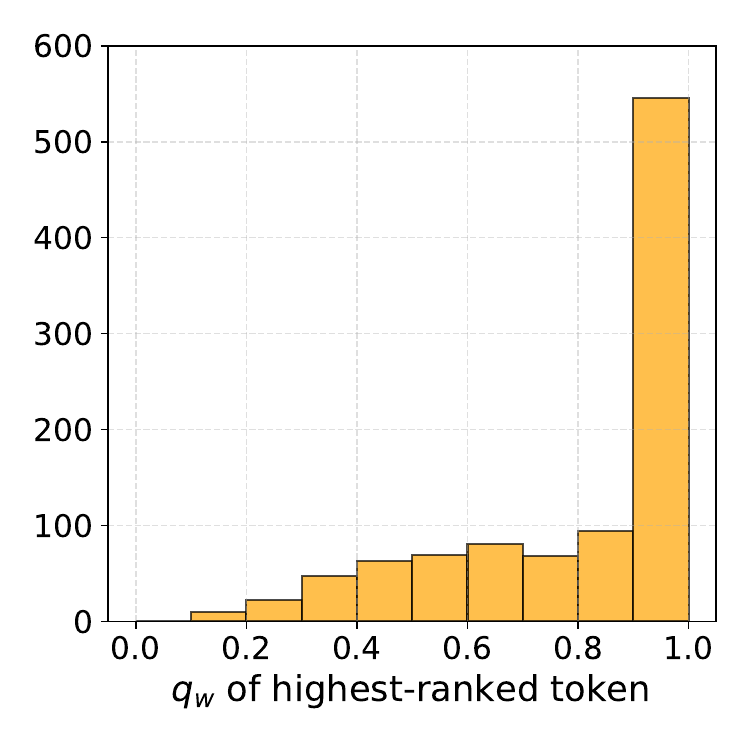}}
    \end{minipage}

    \vspace{0.5em} 
       \begin{minipage}{0.09\linewidth}
        \centering
        \rotatebox{90}{\textbf{SciAbs}}
        \end{minipage}
    \begin{minipage}{0.29\linewidth}
        \centering
        \subfigure[]
        {\includegraphics[width=\linewidth]{fig/dots20.pdf}}
    \end{minipage}
    \hfill
    \begin{minipage}{0.29\linewidth}
        \centering
\subfigure[]        
        {\includegraphics[width=\linewidth]{fig/dots21.pdf}}
    \end{minipage}
    \hfill
    \begin{minipage}{0.29\linewidth}
        \centering
        \subfigure[]
        {\includegraphics[width=\linewidth]{fig/dots22.pdf}}
    \end{minipage}

     \caption{Distributions of the highest-ranked token probabilities
across 1,000 prompts from Wikitext (a-c), XL-Sum (d-f) and SciAbs (g-i) dataset, evaluated at steps 0, 5, and 10. A clear shift toward higher
probability values is observed as AI autophagy unfolds, with a higher collapse in the case of Wikitext.
      }
    \label{fig:other_nexttoken_dists}
\end{figure}

In the main manuscript, Figure~\ref{fig:nexttoken_probs} (d–f) reports the distribution of top-ranked next-token probabilities using SciAbs prompts; here we extend the analysis to other datasets. 
As shown in Figure~\ref{fig:other_nexttoken_dists}, all datasets exhibit a systematic shift toward higher top-token probabilities over simulation steps. 
Model collapse is strongest for Wikitext: by step 5, more than half of the top tokens have a probability that exceeds $0.9$. 
By contrast, SciAbs shows the weakest collapse, with fewer than half tokens with a probability surpassing $0.9$ even at step 10.

\subsection{Evaluation of Gini coefficient and collapsed predictions with prompts of 64 tokens}
\label{sec:app_A3_prompt64token}
\begin{figure}[htb!]
\centering
     \subfigure[]
{\includegraphics[width=0.495\columnwidth]{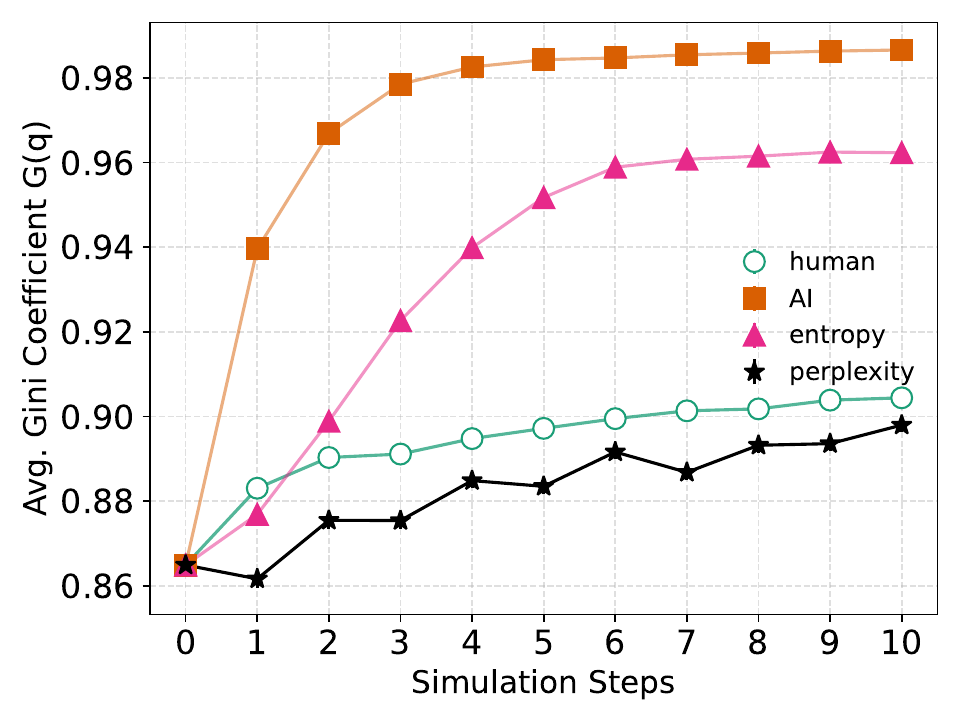}}
    \subfigure[]{\includegraphics[width=0.495\columnwidth]{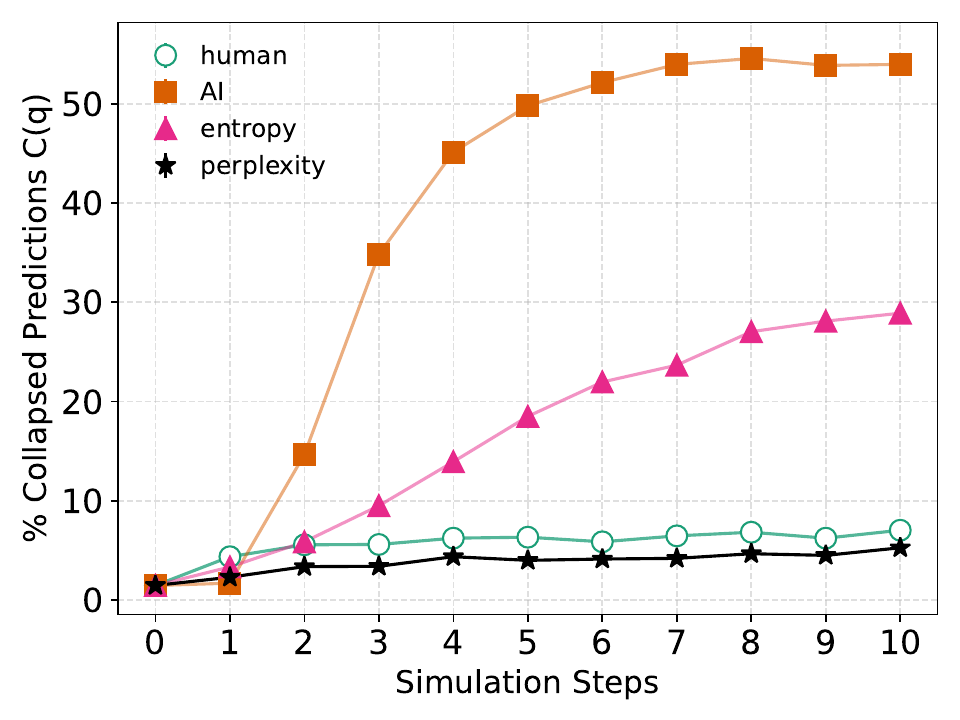}}
\caption{Effects of the perplexity-based filtering strategy on model collapse evaluated on 64 tokens of prompts in next-token predictions.}

\label{fig:mitigation_results_64token}
\end{figure}

In Figure~\ref{fig:mitigation_results} of the main manuscript, we evaluated fine-tuning scenarios using next-token–based metrics with prompts of length $k=32$. Here, we replicate the analysis with $k=64$. 
Figure \ref{fig:mitigation_results_64token} shows that, also for $k=64$, the perplexity-based filtering strategy again achieves the lowest degree of model collapse across all metrics.

\subsection{Mitigation of model collapse using different models}
\label{sec:app_A4_allmodels}

In the main manuscript, all results are reported for Llama2-7B. Here we extend the analysis to Llama3-8B, Mistral-7B, and Qwen3-0.6B, evaluating the same fine-tuning scenarios with our model-collapse metrics. The patterns observed for Llama2-7B generalize: both additional models exhibit the same qualitative trends, confirming the robustness of our findings.

\begin{figure}[!htb]
    \centering
    \vspace{0.1em}

\begin{minipage}{0.495\linewidth}
    \centering
    \includegraphics[width=\linewidth]{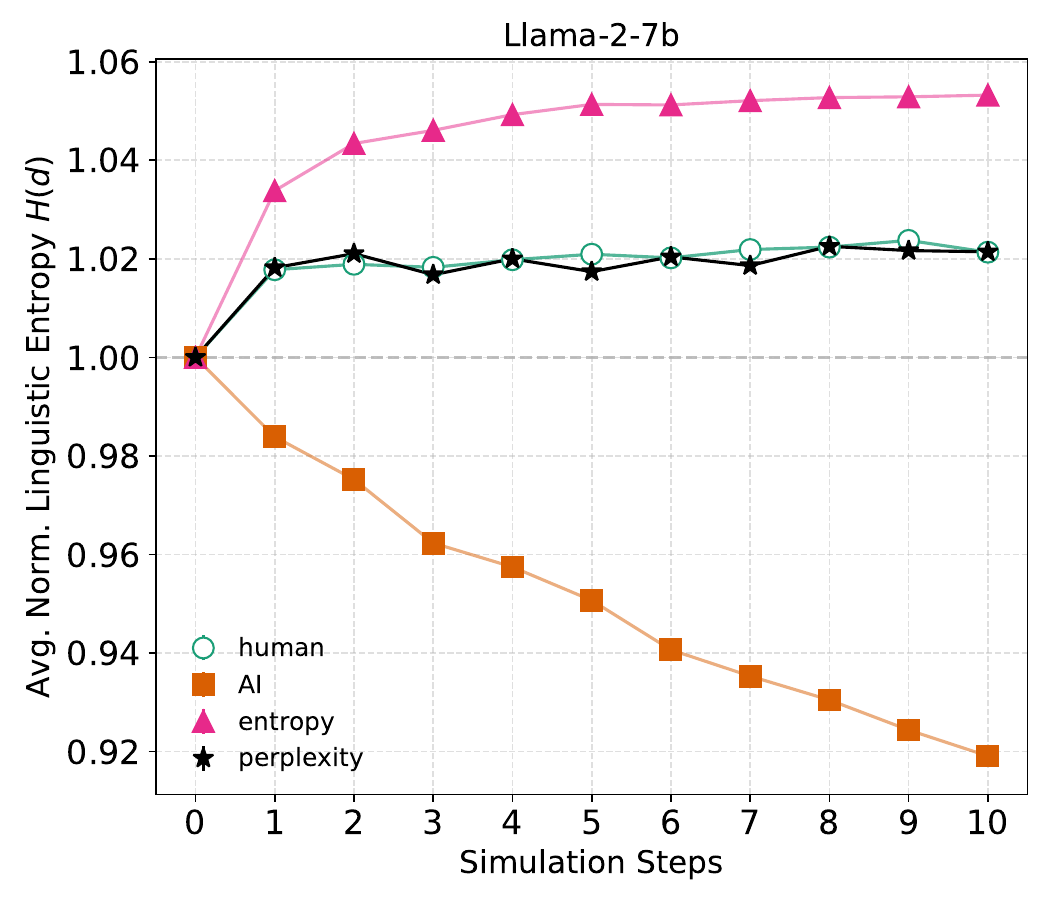}
\end{minipage}
\hfill
\begin{minipage}{0.495\linewidth}
    \centering
    \includegraphics[width=\linewidth]{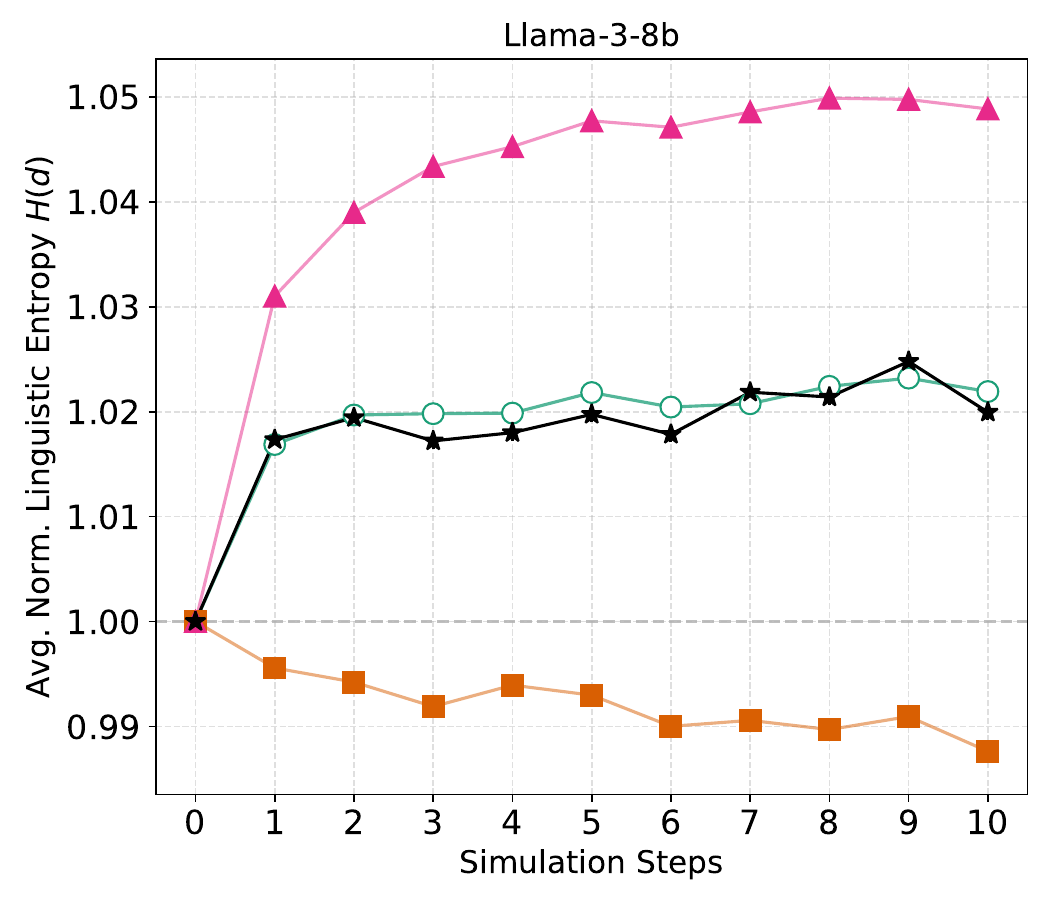}
\end{minipage}

\vspace{0.1em}

\begin{minipage}{0.495\linewidth}
    \centering
    \includegraphics[width=\linewidth]{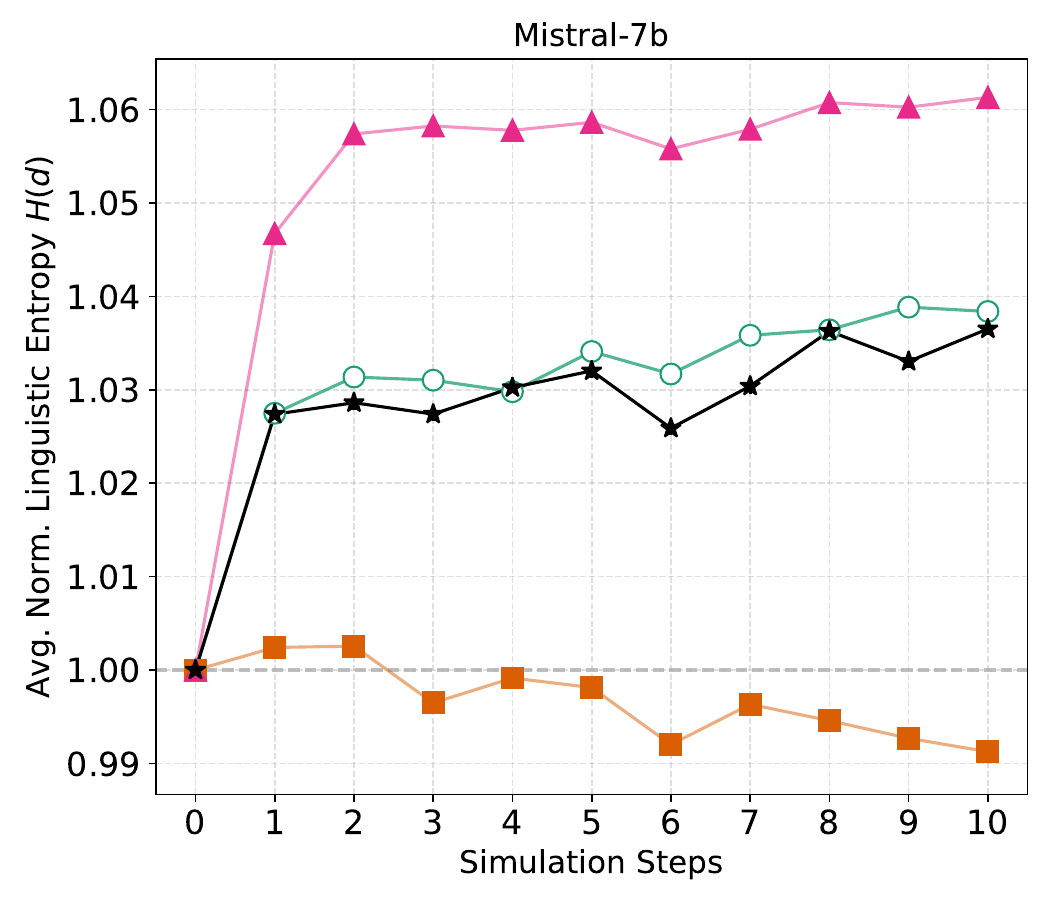}
\end{minipage}
\hfill
\begin{minipage}{0.495\linewidth}
    \centering
    \includegraphics[width=\linewidth]{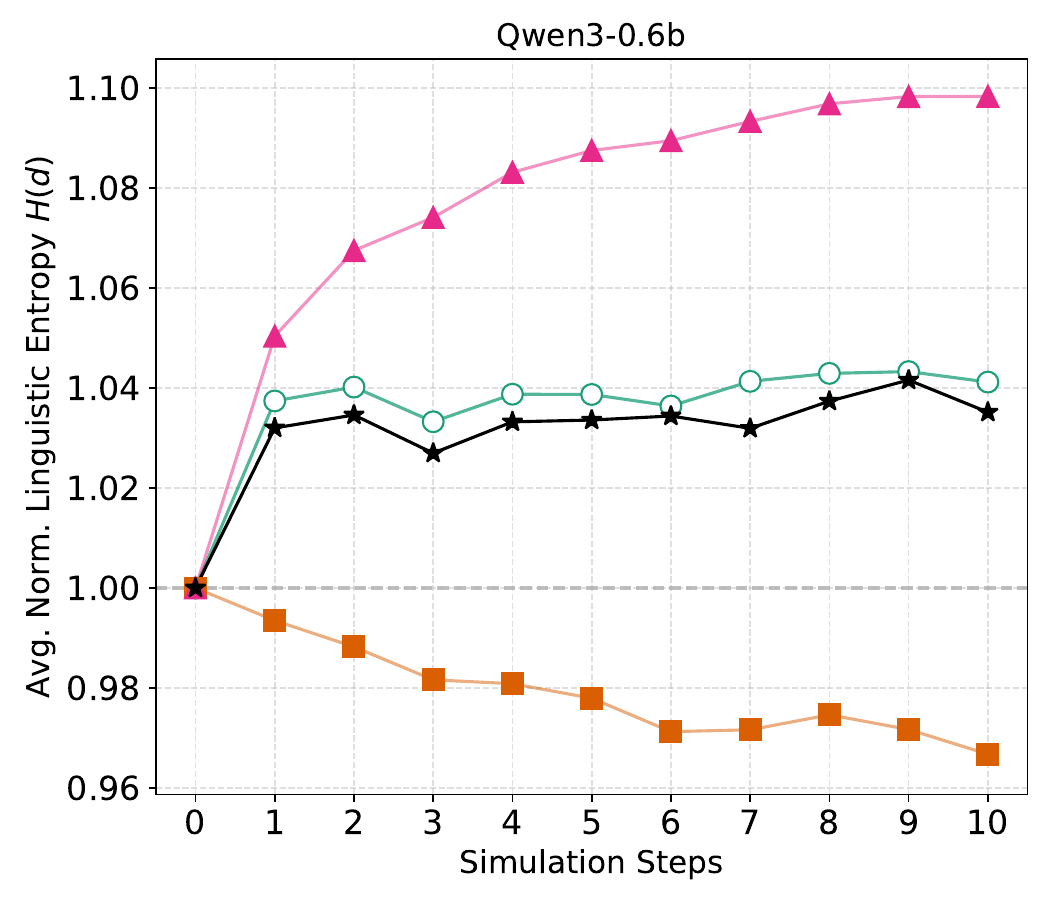}
\end{minipage}

\caption{Linguistic entropy for all fine-tuning scenarios across different models (Llama-2-7b, Llama-3-8b, Mistral-7b and Qwen3-0.6B).}
\label{fig:mitigation_allmodels_entropy}
\end{figure}

\begin{figure}[!htb]
    \centering
    \vspace{0.1em}

\begin{minipage}{0.495\linewidth}
    \centering
    \includegraphics[width=\linewidth]{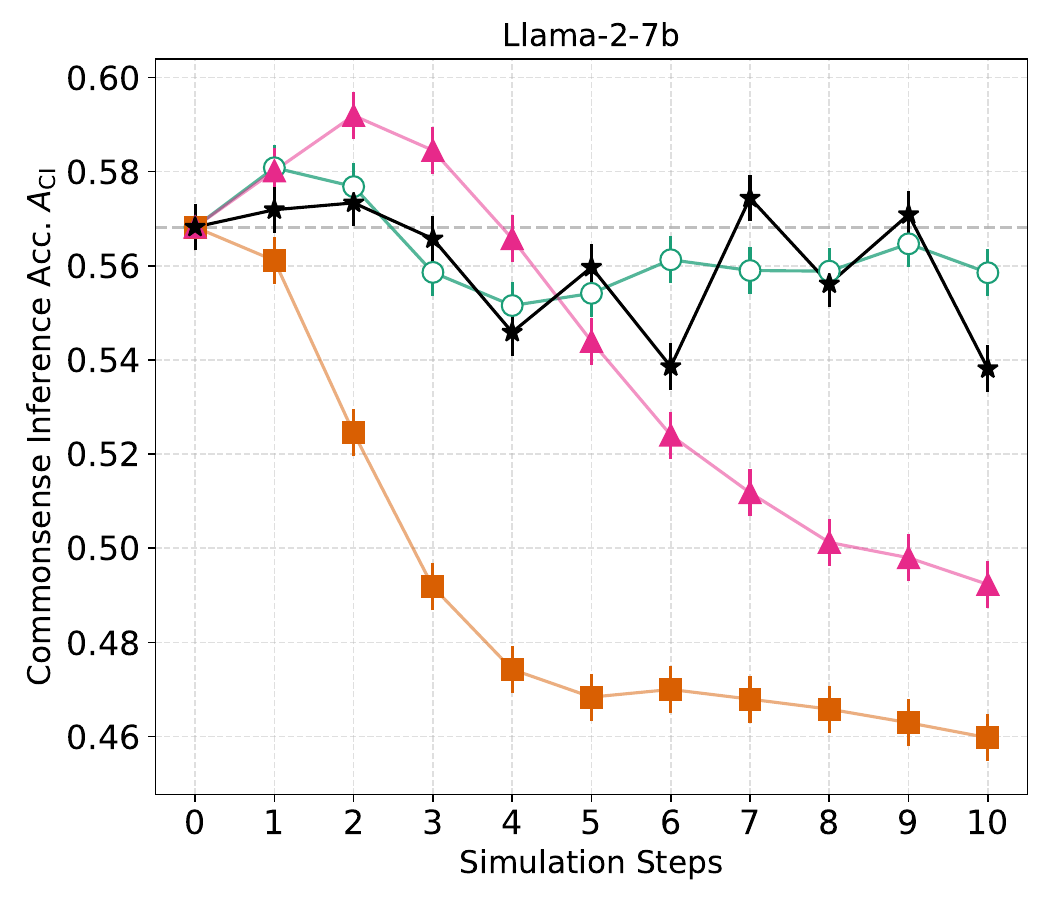}
\end{minipage}
\hfill
\begin{minipage}{0.495\linewidth}
    \centering
    \includegraphics[width=\linewidth]{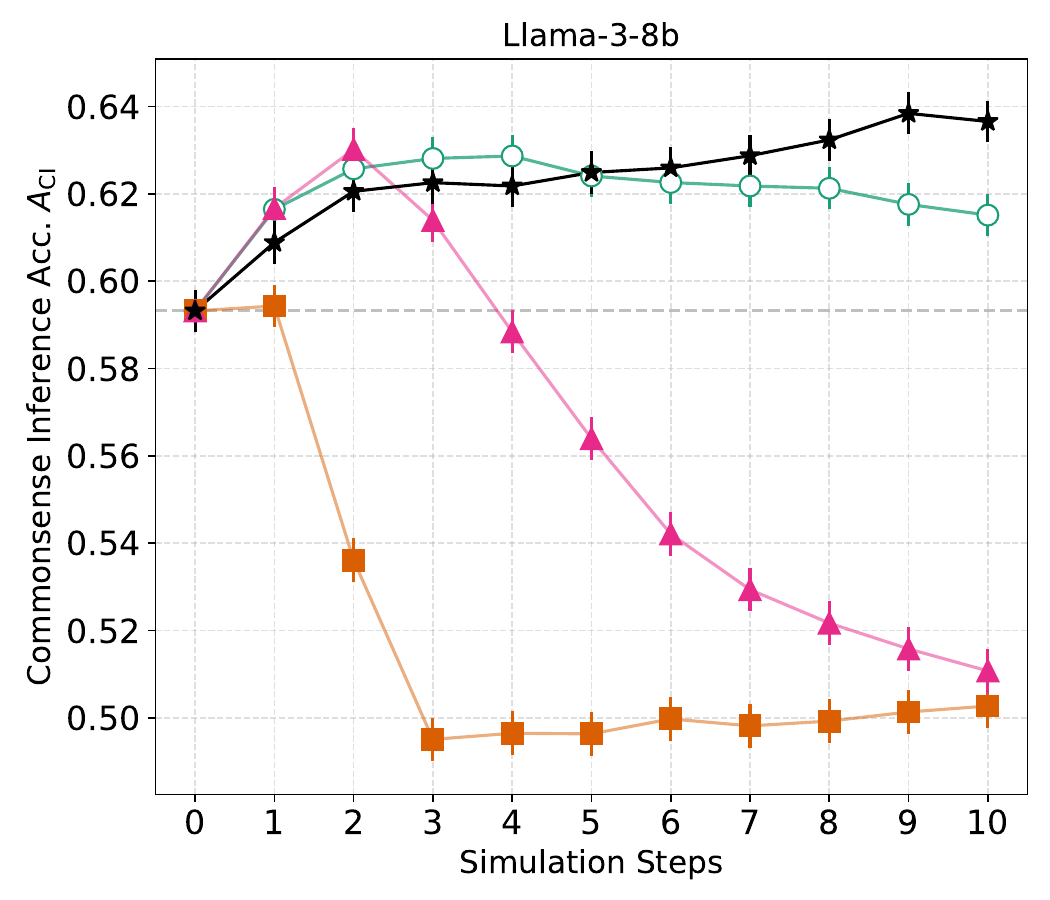}
\end{minipage}

\vspace{0.1em}

\begin{minipage}{0.495\linewidth}
    \centering
    \includegraphics[width=\linewidth]{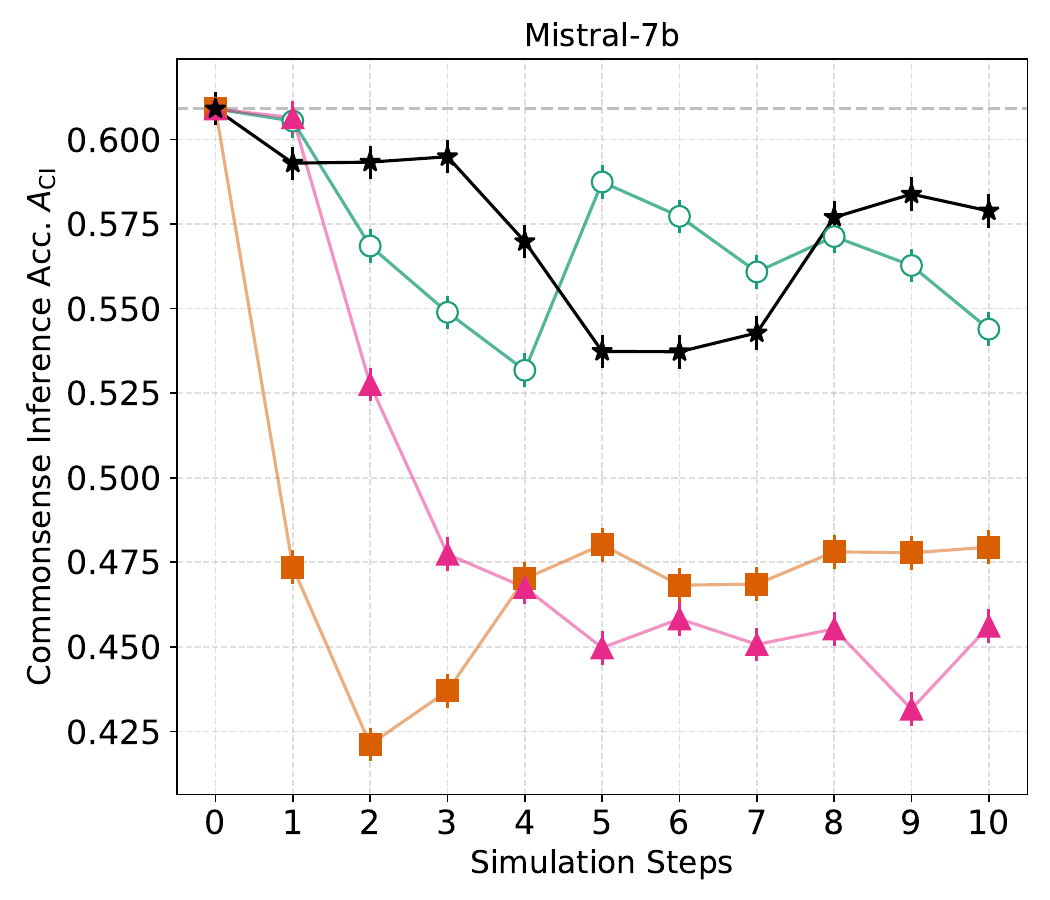}
\end{minipage}
\hfill
\begin{minipage}{0.495\linewidth}
    \centering
    \includegraphics[width=\linewidth]{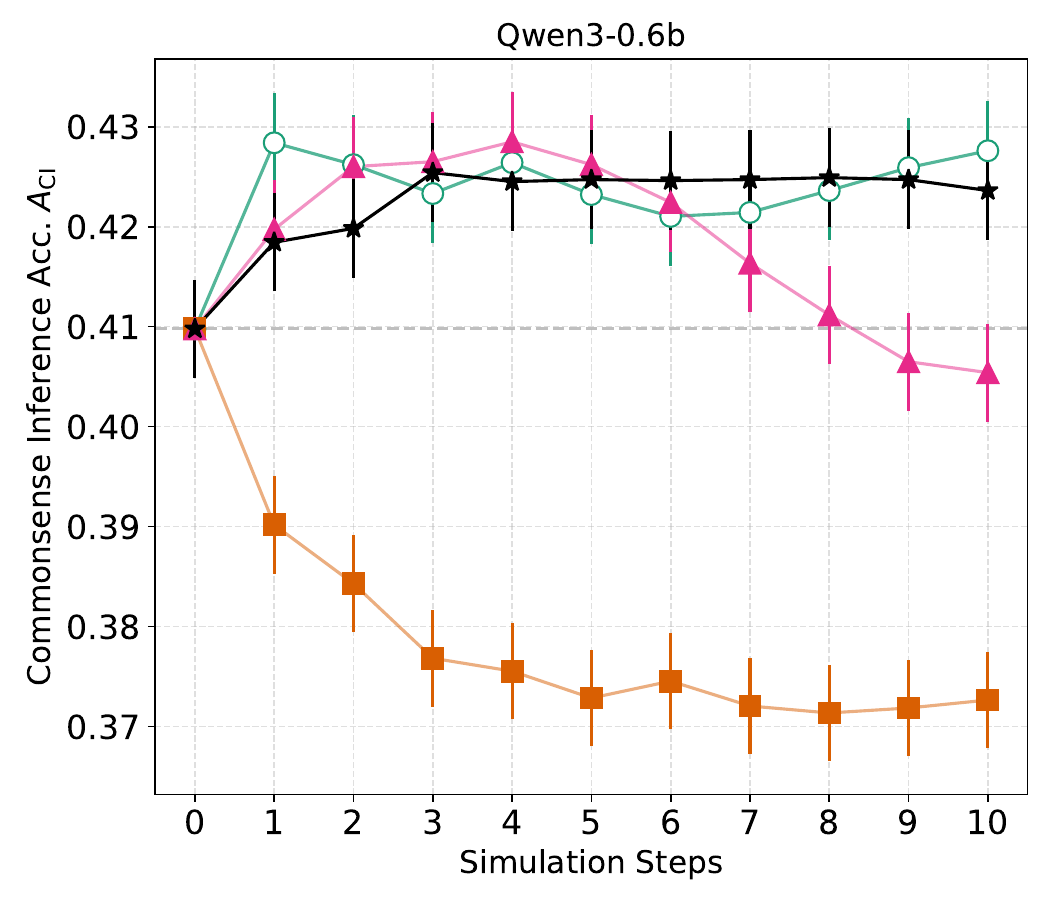}
\end{minipage}

\caption{Commonsense inference accuracy for all fine-tuning scenarios across different models (Llama-2-7b, Llama-3-8b, Mistral-7b and Qwen3-0.6B).}
\label{fig:mitigation_allmodels_hellaswag}
\end{figure}

\end{document}